\documentclass[11pt,a4paper]{article}
\usepackage{coevorsi-layout}
\renewcommand{\coevorsiabstractfont}{\fontsize{9.2pt}{11.5pt}\selectfont}
\usepackage{multirow}
\usepackage{wrapfig}
\usepackage{algorithm}
\usepackage{algpseudocode}
\SetSymbolFont{operators}{normal}{OT1}{cmr}{m}{n}
\SetSymbolFont{operators}{bold}{OT1}{cmr}{bx}{n}
\SetMathAlphabet{\mathit}{normal}{OT1}{cmr}{m}{it}
\SetMathAlphabet{\mathit}{bold}{OT1}{cmr}{bx}{it}
\usepackage{fontawesome5}

\definecolor{lopdteal}{HTML}{287866}
\definecolor{highlightblue}{HTML}{E8F0FE}
\newtcolorbox{insightbox}[1]{
  enhanced jigsaw,
  unbreakable,
  colback=lopdteal!4!white,
  colframe=lopdteal!82!black,
  boxrule=0.8pt,
  arc=4pt,
  left=7pt,
  right=7pt,
  top=6pt,
  bottom=4pt,
  before skip=10pt plus 2pt minus 1pt,
  after skip=10pt plus 2pt minus 1pt,
  fontupper=\small,
  title={#1},
  colbacktitle=lopdteal,
  coltitle=white,
  fonttitle=\scriptsize\bfseries,
  attach boxed title to top left={xshift=12pt,yshift=-2.6mm},
  boxed title style={
    colback=lopdteal,
    colframe=lopdteal,
    boxrule=0pt,
    arc=2pt,
    left=4pt,
    right=4pt,
    top=0.5pt,
    bottom=0.5pt
  }
}

\newcommand{\phead}[1]{{\color{lopdteal!85!black}\bfseries #1}}
\newcommand{\pvar}[1]{{\color{lopdteal}\{#1\}}}
\newcommand{\promptchip}[1]{%
  \node[anchor=west, fill=lopdteal, text=white, rounded corners=2pt,
        inner xsep=4pt, inner ysep=1.6pt, font=\scriptsize\sffamily\bfseries]
    at ([xshift=10pt]frame.north west) {#1};}
\newcommand{\prompttag}[1]{%
  \node[anchor=south east, inner sep=3pt, text=lopdteal!80!black,
        font=\sffamily\fontsize{5.6}{6}\selectfont\itshape]
    at (frame.south east) {#1};}
\newtcolorbox{promptbox}[3][]{
  enhanced,
  breakable,
  frame hidden,
  interior hidden,
  underlay={
    \path[draw=lopdteal!75!black, line width=0.6pt, fill=lopdteal!3!white,
          rounded corners=3pt]
      (frame.south west) rectangle (frame.north east);
  },
  overlay unbroken={\promptchip{#2}\prompttag{#3}},
  overlay first={\promptchip{#2}},
  overlay last={\prompttag{#3}},
  pad before break=4pt,
  pad after break=4pt,
  left=6pt,
  right=6pt,
  top=9pt,
  bottom=11pt,
  before skip=12pt,
  after skip=8pt,
  fontupper=\ttfamily\fontsize{8.3}{10.7}\selectfont\raggedright,
  parbox=false,
  #1
}
\newlist{plist}{itemize}{1}
\setlist[plist]{label={-},leftmargin=1.1em,labelsep=0.4em,nosep,before=\vspace{1pt},after=\vspace{2pt}}
\newlist{penum}{enumerate}{1}
\setlist[penum]{label={\arabic*.},leftmargin=1.4em,labelsep=0.4em,nosep,before=\vspace{1pt},after=\vspace{2pt}}

\usepackage[nameinlink,noabbrev]{cleveref}
\crefformat{section}{\S#2#1#3}
\Crefformat{section}{\S#2#1#3}
\crefname{appendix}{appendix}{appendices}
\Crefname{appendix}{Appendix}{Appendices}
\crefname{subappendix}{appendix}{appendices}
\Crefname{subappendix}{Appendix}{Appendices}
\crefname{subsubappendix}{appendix}{appendices}
\Crefname{subsubappendix}{Appendix}{Appendices}

\newcommand{\ourmethod}{\texttt{CoEvo-RSI}}
\newcommand{\ghlink}{\url{https://github.com/SuJQ0722/CoEvo-RSI}}
\newcommand{\hflink}{\url{https://huggingface.co/collections/nullptr04/coevo-rsi}}
\newcommand{\projectlinks}{\par\vspace{0.6em}\begingroup
  \raggedright\fontsize{9.2pt}{11.5pt}\selectfont
  \renewcommand{\arraystretch}{1.12}\setlength{\tabcolsep}{0pt}
  \begin{tabular}{@{}l@{\hspace{0.55em}}>{\bfseries\color{tealblue}}l@{\hspace{1.1em}}l@{}}
  \faGithub & GitHub & \ghlink \\
  \raisebox{-0.3em}{\includegraphics[height=1.35em]{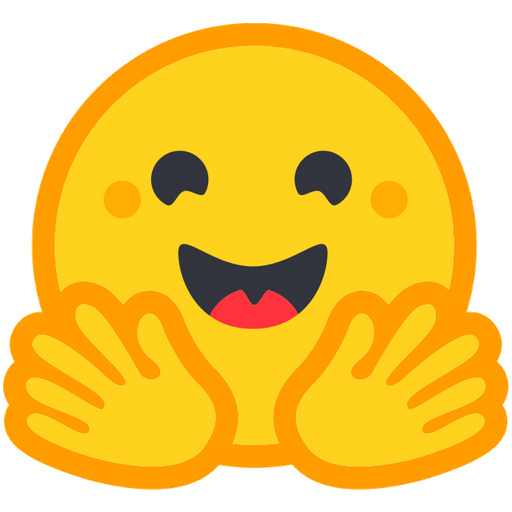}} & Hugging Face & \hflink
  \end{tabular}\par\endgroup}

\hypersetup{
  pdftitle={CoEvo-RSI: When Harnesses Evolve, Which Models Should Follow?},
  pdfauthor={Jiaqi Su, Guibin Zhang, Jie-Jing Shao, Cong Pang, Tiankuo Yao, Ivor Tsang, Haiyan Yin, Shuicheng YAN},
  pdfsubject={Model--harness co-evolution through bidirectional historical inheritance},
  pdfkeywords={CoEvo-RSI, model--harness co-evolution, historical inheritance}
}
\newcommand{\institutionlogoheight}{0.76cm}
\newcommand{\PlaceFirstPageHeader}{\AddToShipoutPictureFG*{%
  \put(\LenToUnit{\dimexpr1in+\oddsidemargin\relax},\LenToUnit{\dimexpr\paperheight-1.593333cm\relax}){%
    \makebox(0,0)[lt]{\makebox[\textwidth][l]{%
      \includegraphics[height=\institutionlogoheight,trim=14bp 5bp 14bp 6bp,clip]{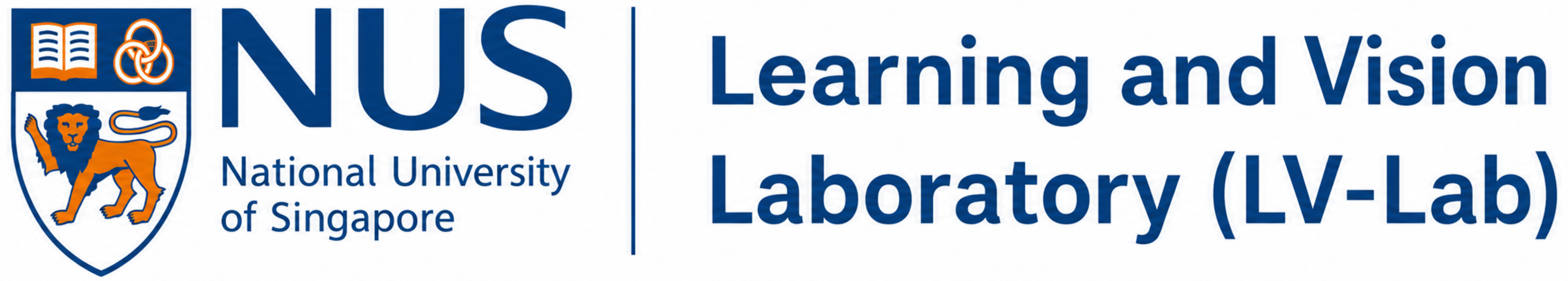}%
      \hfill%
      \includegraphics[height=\institutionlogoheight]{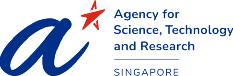}%
      \hspace{0.7cm}%
      \includegraphics[height=\institutionlogoheight]{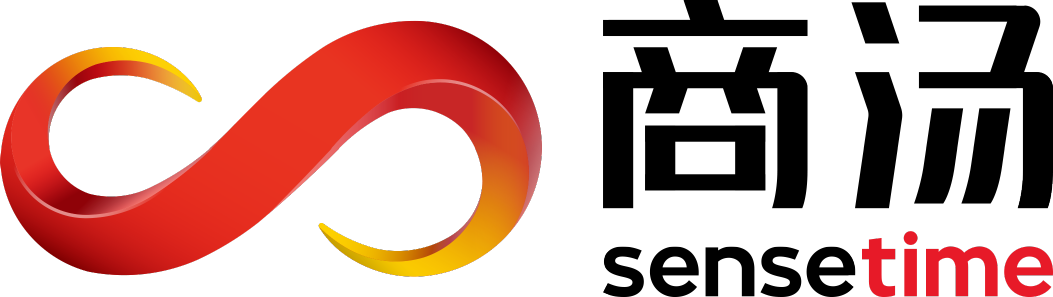}}}}%
  \put(\LenToUnit{\dimexpr1in+\oddsidemargin\relax},\LenToUnit{\dimexpr\paperheight-2.5cm\relax}){%
    \makebox(0,0)[lt]{\rule{\textwidth}{0.4pt}}}%
}}
\title{\vspace*{-0.5cm}%
  \IfFileExists{assets/coevorsi-title.pdf}{%
    \includegraphics[width=0.35\textwidth,trim=0 140bp 0 120bp,clip]{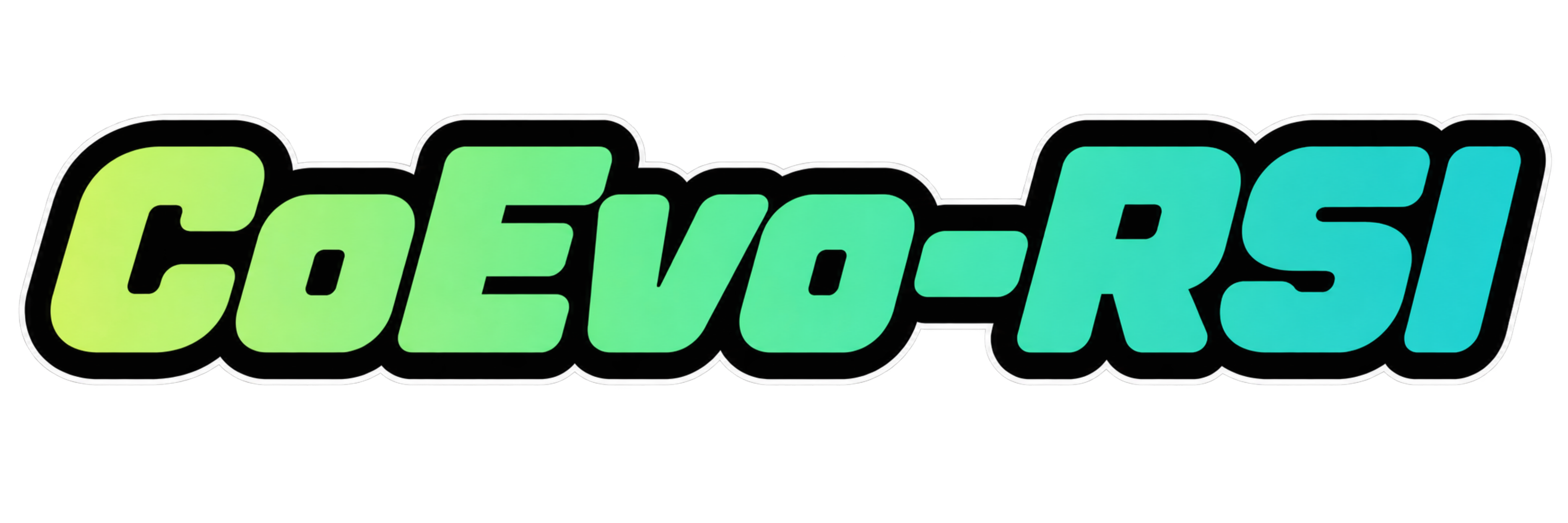}%
  }{\fontsize{30pt}{32pt}\selectfont\bfseries CoEvo-RSI}\\[0.5em]
  \fontsize{15pt}{18pt}\selectfont\textbf{When Harnesses Evolve, Which Models Should Follow?}}
\author{\textbf{Jiaqi Su}, \;\textbf{Guibin Zhang},\;\textbf{Jie-Jing Shao},\;\textbf{Cong Pang}\\[0.15em]
\textbf{Tiankuo Yao},\;\textbf{Ivor Tsang},\;\textbf{Haiyan Yin},\;\textbf{Shuicheng YAN}\\[0.45em]
{\fontsize{9.2pt}{11.4pt}\selectfont National University of Singapore}\\[-0.1em]
{\fontsize{9.2pt}{11.4pt}\selectfont Agency for Science, Technology and Research (A*STAR)}\\[-0.1em]
{\fontsize{9.2pt}{11.4pt}\selectfont SenseTime Group Inc.}}
\date{}

\begin{document}
\PlaceFirstPageHeader
\maketitle
\vspace{-0.2em}

\begin{abstract}
Model--harness co-evolution has emerged as a prominent paradigm for recursive self-improving (RSI) agents, jointly evolving their execution scaffolds and intrinsic capabilities.
We study the persistent yet often overlooked \emph{brain--body mismatch problem} in this paradigm, where an evolving harness (body) demands behaviors its model (brain) has not learned or restricts capabilities it already possesses.
We introduce \ourmethod{}, a RSI framework that advances model--harness co-evolution through agent-driven, bidirectional historical inheritance.
Drawing on a shared evolutionary archive and execution feedback, a meta-agent coordinates harness development from historical designs with model checkpoint selection or merging and subsequent adaptation under the selected harness, aligning inherited capabilities with evolving execution requirements to reduce brain--body mismatch.
Using only supervised fine-tuning (SFT) on its own rejection-sampled trajectories, \ourmethod{} makes Qwen3.5-9B competitive with larger models across diverse vertical applications, including multi-agent failure attribution, role-aware text-to-SQL, and smart-home instruction following.
It surpasses DeepSeek-V4-Pro on TraceElephant ($+14.16$ agent accuracy) and GLM-5.3 on HomeBench VM ($+4.90$ success rate), underscoring the need for brain--body alignment and the potential of bidirectional historical inheritance to advance agent co-evolution.
\projectlinks
\end{abstract}

\par\addvspace{\intextsep}
\noindent\begin{minipage}{\textwidth}
\centering
\vspace{1em}
\includegraphics[width=\textwidth]{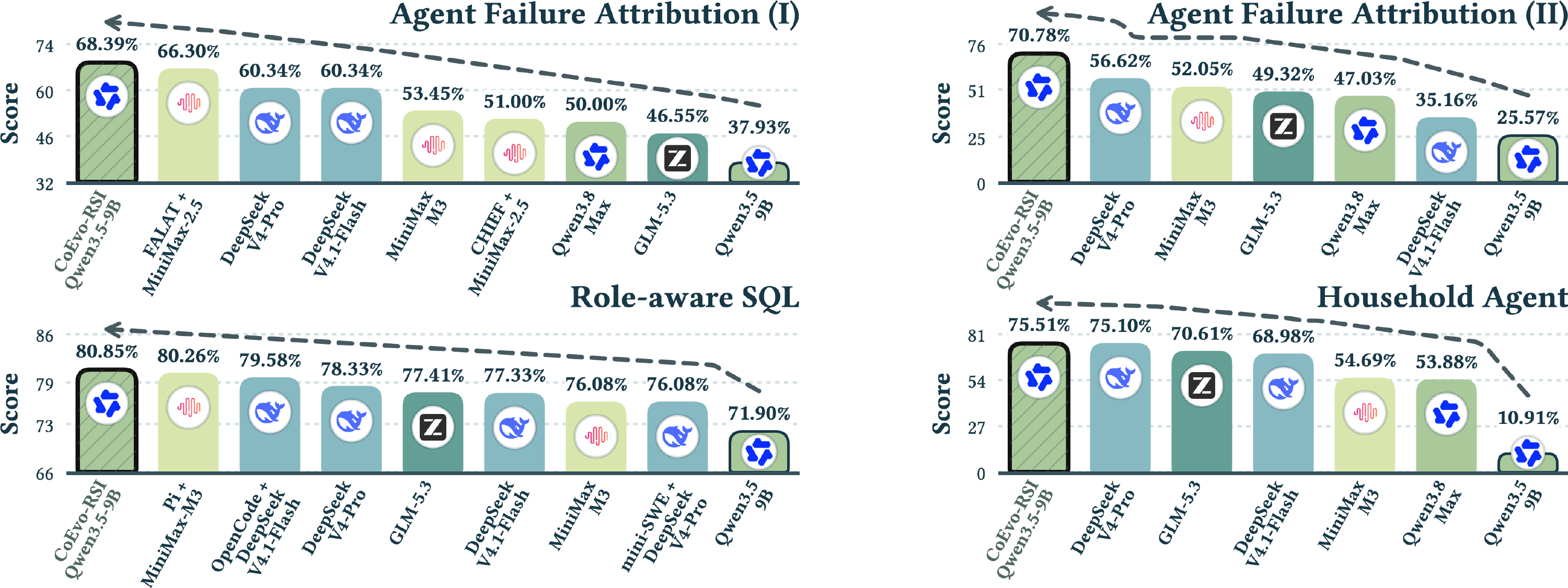}
\captionof{figure}{\textbf{Co-evolution elevates Qwen3.5-9b-based agent across three domains.}
Agent Failure Attribution (I) and (II) report Who accuracy (correct identification of the responsible agent) on Who\&When Hand and TraceElephant, respectively.
Role-aware SQL reports safe execution accuracy (Safe-EX) on RBAC Spider; Household Agent reports success rate on HomeBench's valid multi-device instructions (VM).
All scores are percentages, with higher being better. Hatching marks \ourmethod{}, and the outlined solid Qwen3.5-9B bar is its base model.
Dashed arrows trace the improvement from the base model to \ourmethod{}. }
\label{fig:benchmark-leaderboard}
\end{minipage}\par
\clearpage

\section{Introduction}
\label{sec:intro}
An agent's capabilities depend jointly on its \emph{brain}---the language model that reasons and selects actions---and its \emph{body}---the harness that defines its tool interfaces, action space, memory, and execution workflow~\citep{lou2026autoharness,chen2026harnessx}. Accordingly, progress has followed two complementary directions. Model training improves the agent's learned policy through supervised learning, reinforcement learning, and self-distillation~\citep{xia2026skillrl,yang2026coevolve,zhang2026lopd}. Harness optimization improves the surrounding system with the model fixed, through iterative modification of prompts, tools, and control flow guided by execution feedback~\citep{lou2026autoharness,lin2026ahe}, or population-based evolution that generates, evaluates, and recombines competing harness designs~\citep{lee2026metaharness,zhang2026darwinx}. These directions meet in \textbf{model--harness co-evolution}, where model training and harness optimization jointly adapt an agent to its task domain~\citep{chen2026harnessx,luo2026hase}. Recent systems demonstrate the promise of this approach across tool use, mathematical reasoning, search-based question answering, and chess~\citep{chen2026harnessforge,chen2026coharness,kim2026whale}.

A co-evolution round couples harness evolution with model adaptation. Starting from a model--harness pair $(\mathbf{H}_t,\theta_t)$, execution feedback $\mathbf{F}_t$ guides changes to the harness, whose requirements in turn shape the model's next adaptation~\citep{chen2026coharness}. As either component evolves, the harness's execution requirements and the model's learned capabilities can drift apart. In this setting, we first identify and formalize the \textbf{brain--body mismatch problem}:

\begin{insightbox}{Problem Definition}
\emph{During model--harness co-evolution, a pair $(\mathbf{H}_t,\theta_t)$ exhibits \textbf{brain--body mismatch} when the behaviors required by $\mathbf{H}_t$ diverge from those learned by $\theta_t$, or when $\mathbf{H}_t$ restricts the use of capabilities already present in $\theta_t$. The problem is to maintain their compatibility through successive rounds of harness-model co-adaptation.}
\end{insightbox}

\begin{wrapfigure}{r}{0.6\columnwidth}
\centering
\includegraphics[width=\linewidth]{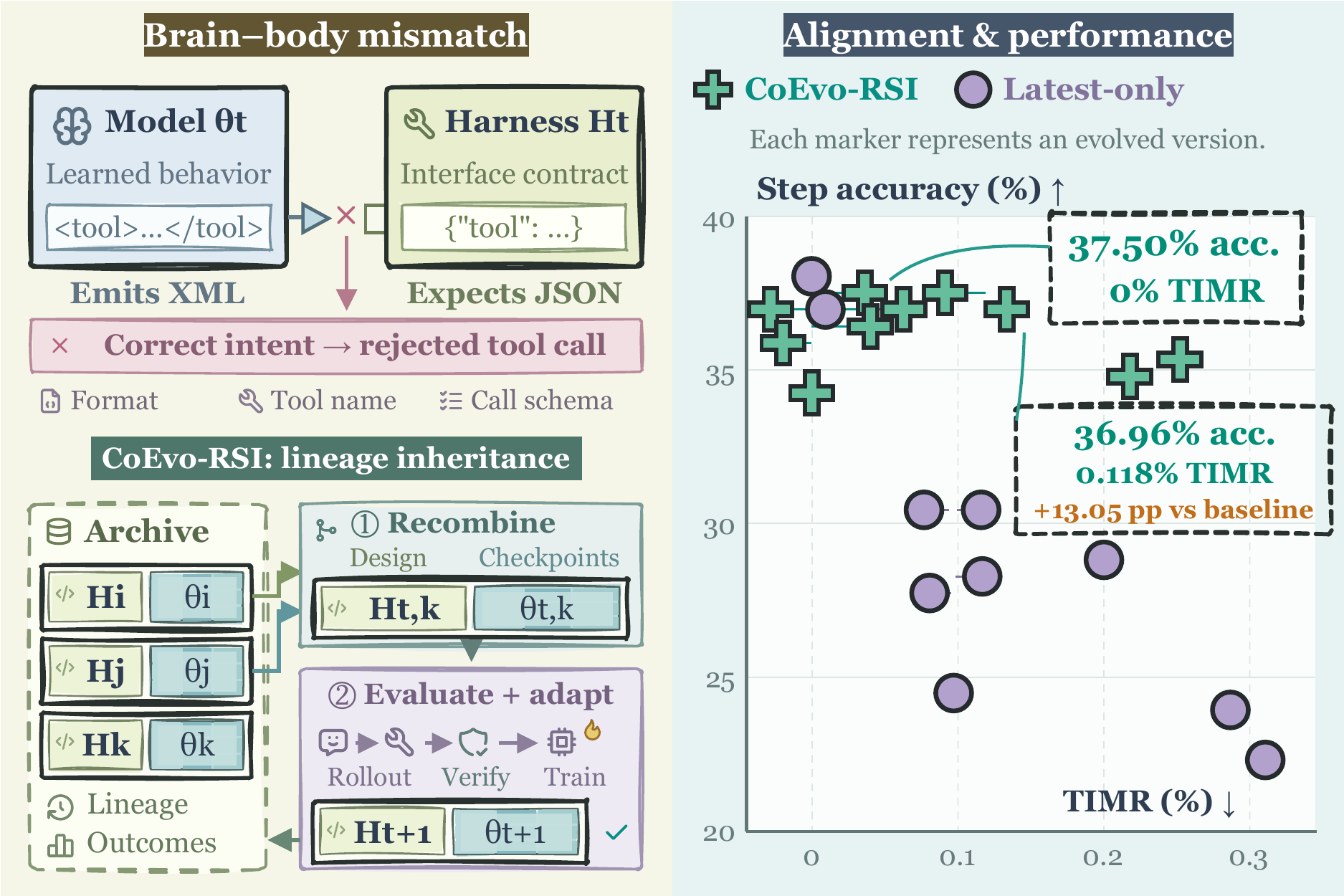}
\caption{\textbf{Brain--body alignment in model--harness co-evolution.}
\textbf{Left:} Historical harnesses and checkpoints are recombined and adapted to resolve interface mismatch.
\textbf{Right:} Step accuracy versus Turn-level Interface Mismatch Rate (TIMR; \Cref{sec:coevolution}) for ten evolved versions per method, from a three-round run evaluated on the Who\&When test benchmark.}
\label{fig:coevox-teaser}

\end{wrapfigure}

In plain terms, the model may struggle to operate its harness, and the harness may underuse its model's abilities (\Cref{fig:coevox-teaser}). Consider two concrete examples. \ding{182}~\emph{Interface mismatch.} The model emits XML tool calls, while the evolved harness expects JSON, causing otherwise correct calls to be rejected. \ding{183}~\emph{Capability mismatch.} The harness introduces an \emph{active context management tool} for deciding what information to retain, summarize, or discard. A model unfamiliar with these operations may use it poorly; a historical checkpoint $\theta_i$ already possesses the relevant capabilities and provides a suitable starting point for adaptation. On the harness side, a historical design $\mathbf{H}_j$ can supply an execution workflow suited to the model's capabilities. These examples motivate selecting predecessors whose inherited behaviors and structures support further evolution.

To address this challenge, we propose \ourmethod{}, a framework for \textbf{model--harness co-evolution through bidirectional historical inheritance}. Its \textbf{evolutionary archive} $\mathbf{A}_{<t}$ records earlier harnesses, model checkpoints, their development histories, and execution outcomes. Each round begins with feedback from the current pair. A Curator organizes this evidence, and harness editors browse the archive, choose their own historical starting designs, and develop offspring harnesses. For each offspring, the framework selects or merges historical checkpoints to form candidate model--harness pairs. Evaluation identifies a promising pair for model adaptation under its harness, and validation determines which pair advances to the next round. The resulting designs, checkpoints, and outcomes extend the archive. This \textbf{agent-driven inheritance} connects the two sides of evolution: harness design draws on accumulated structural experience, while model initialization draws on capabilities suited to the emerging harness. Their coordination makes \emph{brain--body alignment} the organizing principle of the evolutionary process.
Our contributions are:

\begin{itemize}[leftmargin=*, itemsep=0pt, parsep=0pt]
    \item \textbf{Problem Formalization.}
    We formalize brain--body mismatch in model--harness co-evolution and introduce the turn-level interface mismatch rate (TIMR) as an inexpensive diagnostic signal for observable interface violations.
    \item \textbf{Method Design.}
    We propose \ourmethod{}, which combines an evolutionary archive with bidirectional historical inheritance. Harness editors select historical starting designs, and model inheritance selects or merges checkpoints to support adaptation under the evolved harness.
    \item \textbf{Empirical Evaluation.}
    We evaluate \ourmethod{} across multi-agent failure attribution, home care, and SQL tasks, with co-evolved Qwen3.5-9B outperforming Qwen3.8-Max on Who\&When Pro ($+4.40$ agent acc.) and DeepSeek-V4-Pro on TraceElephant ($+14.16$ agent acc.).
\end{itemize}

\section{Related Work}

Self-evolving agents improve through changes to external execution systems, learned model parameters, or both~\citep{gao2025selfevolving,guo2026harnesssurvey,wang2026diving}. We organize the literature around these optimization targets and the reuse of historical components that connects them.

\paragraph{Recursive Harness Improvement.}
Early agent systems established how external organization can improve a model's behavior: ReAct interleaves reasoning and actions, Reflexion retains verbal feedback across trials, and Voyager accumulates reusable executable skills~\citep{yao2023react,shinn2023reflexion,wang2023voyager}. Agent--computer interface design further shows how the operations exposed to a model shape its effectiveness~\citep{yang2024sweagent}. Subsequent work turns these hand-designed mechanisms into optimization targets, evolving context~\citep{zhang2025ace}, memory~\citep{zhang2025memevolve,zhang2026memskill,pan2026mstar}, planning~\citep{liu2026todoevolve}, and retrieval~\citep{lee2026selfindex}. At the system level, harness search jointly modifies prompts, tools, and control flow~\citep{lee2026rhi,lee2026metaharness,lin2026ahe,lou2026autoharness,zhang2026selfharness,zhang2026hyperagents}. AutoHarness synthesizes executable support from environmental feedback, while Meta-Harness gives its proposer access to earlier source code, scores, and execution traces~\citep{lou2026autoharness,lee2026metaharness}. Such access preserves evidence that a single scalar score or compressed reflection can omit. Online adaptation extends this process to deployment streams and test-time feedback~\citep{liu2026adaptiveharness,nie2026tthe,pan2026rho}; population search and reusable repositories retain alternative designs for subsequent exploration~\citep{zhang2026darwinx,nie2026evolvenet,luo2026harnessbank}. The Darwin G\"odel Machine similarly grows an archive of self-modifying agents, allowing earlier branches to become stepping stones for later improvements~\citep{zhang2025dgm}. Together, these approaches expand the search target from isolated modules to complete execution systems. They also motivate preserving design diversity and execution history, which \ourmethod{} connects to a corresponding archive of model checkpoints.

\paragraph{Recursive Model Improvement.}
Model self-improvement converts generated experience into persistent changes in the policy. Three learning loops are particularly relevant. \textbf{(i) Curricula} adapt the source and difficulty of training problems to the learner, including self-play, tool-oriented task generation, and search-agent training without externally supplied demonstrations~\citep{zhao2025absolutezero,acikgoz2026toolr0,sundaram2026soar,yang2026ttcs,yang2026coevolve,yue2026drzero}. The resulting feedback loop changes which experiences the model encounters as its competence develops. \textbf{(ii) Supervision} turns demonstrations, privileged feedback, or the model's own improved behavior into self-distillation targets~\citep{zhao2026opsd,shenfeld2026sdft,hubotter2026sdpo,zhang2026lopd,lu2026sdar}. On-policy variants address the learner's current behavior, making the distribution of generated trajectories central to what can be learned. \textbf{(iii) Experience reuse} extracts reusable skills and transfers their guidance into model behavior~\citep{xia2026skillrl,zhang2026opdevolver,lu2026skill0,wu2026seed,yang2026opid}. For example, SkillRL jointly evolves a hierarchical skill library and the agent policy, linking experience abstraction with subsequent reinforcement learning~\citep{xia2026skillrl}. These mechanisms improve the agent's intrinsic capabilities, while the harness determines which tools, observations, and execution constraints appear in its training trajectories. This dependence motivates choosing model initialization and adaptation in relation to the harness that will actually execute the policy.

\paragraph{Model--Harness Co-Evolution.}
Joint adaptation closes the loop between the environment presented to a policy and the behavior learned under that environment. One line couples \textbf{(i) context and weights} through prompt optimization, continual adaptation, and the transfer of contextual guidance into parameters~\citep{lu2026p2o,tiwari2026fst,tu2026ucob,zhao2026ophsd}. A second line couples \textbf{(ii) executable harnesses and policies}, extending adaptation to the tools, memory, middleware, and control flow used during rollout~\citep{chen2026harnessforge,chen2026harnessx,chen2026coharness,kim2026whale,qi2026vace,qu2026qwenplanneragent,hebbar2026sia,karten2026continualharness,mao2026safeevolve,taolive2026hat,yu2026onpolicycorrection}. HarnessForge combines fault-guided harness tailoring with harness-conditioned policy alignment~\citep{chen2026harnessforge}. Co-Harness alternates validated harness patches with fine-tuning on trajectories generated by the improved scaffold~\citep{chen2026coharness}, while WHALE alternates rejection-sampling fine-tuning with executable harness search~\citep{kim2026whale}. A third line brings \textbf{(iii) execution and editing} into a shared learning process, allowing policies to participate in the evolution of their own support~\citep{luo2026hase,shi2026skill1}. These methods establish the value of adapting both sides and expose a common dependency: a harness update changes the execution requirements, and a model update changes which harnesses are effective. \ourmethod{} makes \emph{bidirectional historical inheritance} an explicit decision within this loop. Harness editors choose historical starting designs, and model inheritance independently selects or merges archived checkpoints for the resulting requirements. This enables recovery of useful earlier capabilities without tying model initialization to the latest generation or the harness's parent branch. \Cref{app:evolution-comparison} compares the inheritance procedures of representative frameworks.

\section{Method}
\label{sec:method}

\begin{figure}[!t]
\centering
\includegraphics[width=\textwidth]{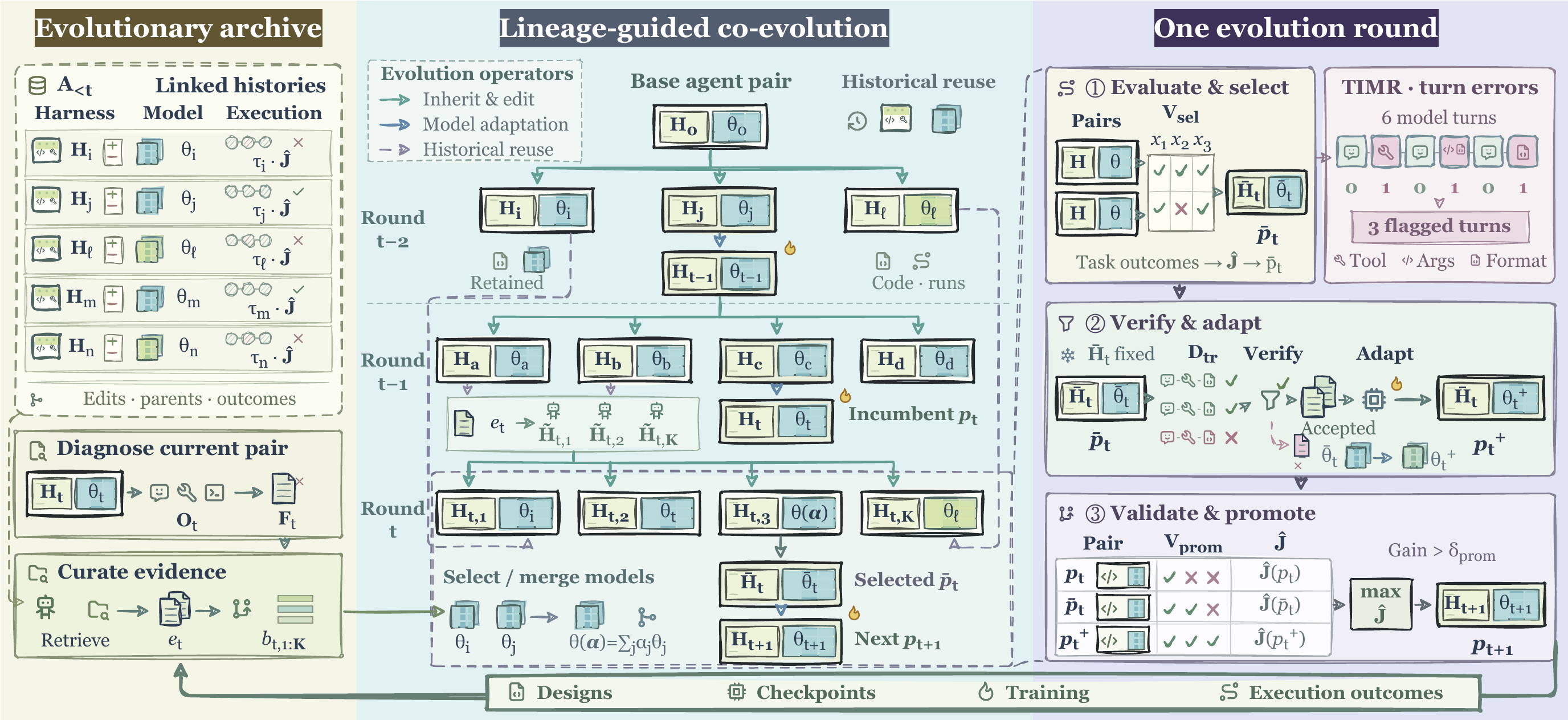}
\caption{\textbf{Overview of \ourmethod{}.}
\textbf{Left:} The Curator draws on archived designs, checkpoints, and outcomes to guide evolution.
\textbf{Center:} Harness editors choose historical harness bases; checkpoint selection or merging supplies the paired models.
\textbf{Right:} Each round evaluates candidate pairs, tracks TIMR, adapts the selected model to its harness, and promotes improvements by validation performance. Outcomes return to the archive.}
\label{fig:coevox-framework}

\end{figure}

\ourmethod{} (\Cref{fig:coevox-framework}) realizes model--harness co-evolution through \textbf{bidirectional historical inheritance}. An autonomous meta-agent coordinates two complementary decisions: harness editors select and develop historical designs, and model inheritance selects or merges checkpoints for the resulting execution requirements. Evaluation, model adaptation, and promotion turn these inherited components into the next agent, whose outcomes enrich the evolutionary archive.

\suppressfloats[t]

\subsection{Model--Harness Co-Evolution}
\label{sec:coevolution}
An executable agent is a pair $p=(\mathbf{H},\theta)$, where the harness $\mathbf{H}$ specifies prompts, tools, state management, and control flow, and $\theta$ denotes the task model's parameters. Constraints $\mathbf{C}$ fix the environment, verifier, and admissible interfaces. Let $\rho$ be the task distribution and $\mathbf{P}_{\theta,\mathbf{H}}(\cdot\mid x;\mathbf{C})$ the distribution of execution trajectories for task $x$. A trajectory $\tau$ contains model responses, tool interactions, and the final output; its task score $s(x,\tau)$ is assigned by the fixed evaluator. The objective is
\begin{equation}
\mathbf{J}(\mathbf{H},\theta)
=\mathbf{E}_{x\sim\rho}\mathbf{E}_{\tau\sim\mathbf{P}_{\theta,\mathbf{H}}(\cdot\mid x;\mathbf{C})}[s(x,\tau)],
\label{eq:lineage-objective}
\end{equation}
where $\mathbf{E}$ denotes expectation. For a task set $\mathbf{D}$ and $n_{\mathrm{eval}}$ executions per task, let $\tau_{x,u}$ be the trajectory from execution $u$. We estimate this objective by
\begin{equation}
\widehat{\mathbf{J}}_{\mathbf{D}}(\mathbf{H},\theta)
=\frac{1}{|\mathbf{D}|n_{\mathrm{eval}}}
\sum_{x\in\mathbf{D}}\sum_{u=1}^{n_{\mathrm{eval}}}s(x,\tau_{x,u}).
\label{eq:lineage-estimate}
\end{equation}
For $p=(\mathbf{H},\theta)$, we also write $\widehat{\mathbf{J}}_{\mathbf{D}}(p)$ for this estimate.

Each generation develops and adapts model--harness pairs through two inheritance steps. A meta-agent with fixed parameters $\phi$ and role policies $\pi_\phi^{\mathrm{role}}$ operates on the incumbent $p_t=(\mathbf{H}_t,\theta_t)$ and an archive $\mathbf{A}_{<t}$ of earlier harnesses, checkpoints, and execution histories, including the seed pair. First, the incumbent's execution feedback $\mathbf{F}_t$ guides the generation of $\mathbf{K}$ offspring harnesses $\mathbf{H}_{t,1:\mathbf{K}}$ and shared evidence $e_t$:
\begin{equation}
(\mathbf{H}_{t,1:\mathbf{K}},e_t)
=\operatorname{Inherit}_{\phi}^{\mathrm{h}}(p_t,\mathbf{A}_{<t},\mathbf{F}_t),
\label{eq:overview-harness-inheritance}
\end{equation}
where $e_t$ links historical designs to their models and outcomes. Second, model inheritance pairs these offspring with selected or merged checkpoints:
\begin{equation}
\mathbf{Q}_t^{\mathrm{sel}}
=\operatorname{Inherit}_{\phi}^{\mathrm{m}}(\mathbf{H}_{t,1:\mathbf{K}},\mathbf{A}_{<t},e_t),
\label{eq:overview-model-inheritance}
\end{equation}
where $\mathbf{Q}_t^{\mathrm{sel}}$ is the candidate pair set. Evaluation selects $\bar p_t\in\mathbf{Q}_t^{\mathrm{sel}}$, adaptation produces $p_t^+$, and validation chooses $p_{t+1}$ from $p_t$, $\bar p_t$, and $p_t^+$. Completed outcomes extend the archive to $\mathbf{A}_{<t+1}$.

\paragraph{Diagnosing interface mismatch.}
We introduce the \emph{Turn-level Interface Mismatch Rate} (TIMR) as a diagnostic signal for interface mismatch.
TIMR captures observable interface violations; broader capability alignment calls for complementary evaluation, as discussed in \Cref{app:timr}.
We do not claim that TIMR fully characterizes brain--body mismatch; it captures one aspect while serving as a cheap-to-verify and empirically useful metric. For a pair's evaluated trajectories $\mathbf{Z}(\mathbf{H},\theta)$, let $n_{\mathrm{lm}}(\tau)$ count model turns and $v_{\tau,u}\in\{0,1\}$ flag calls to unavailable tools, interface-contract violations, or malformed outputs at turn $u$. We define
\begin{equation}
\mathbf{TIMR}(\mathbf{H},\theta)=
\frac{\sum_{\tau\in\mathbf{Z}(\mathbf{H},\theta)}\sum_{u=1}^{n_{\mathrm{lm}}(\tau)}v_{\tau,u}}
{\sum_{\tau\in\mathbf{Z}(\mathbf{H},\theta)}n_{\mathrm{lm}}(\tau)},
\label{eq:timr}
\end{equation}
where each violating turn counts once, even if later recovered; lower TIMR indicates fewer interface violations per model turn.

\subsection{Harness Inheritance}
\label{sec:harness-inheritance}
We instantiate \Cref{eq:overview-harness-inheritance} through evidence curation and historical base selection, linking offspring design to recorded model--harness compatibility.

\paragraph{Evidence curation.}
To find designs that address the incumbent's failures, we diagnose observations $\mathbf{O}_t$ from $p_t$ on the validation set $\mathbf{V}$, obtaining $\mathbf{F}_t=\operatorname{diagnose}(p_t,\mathbf{O}_t)$. The Curator uses this feedback to retrieve relevant history. At step $q$, it chooses action $u_{t,q}$ from its inspection history $r_{t,q}^{\mathrm{cur}}$:
\begin{equation}
u_{t,q}\sim\pi_\phi^{\mathrm{ret}}\bigl(\cdot\mid\mathbf{H}_t,\mathbf{F}_t,\operatorname{index}(\mathbf{A}_{<t}),r_{t,q}^{\mathrm{cur}}\bigr),
\label{eq:lineage-retrieval}
\end{equation}
where $\operatorname{index}$ exposes archive metadata. Each action and its result $\operatorname{read}(\mathbf{A}_{<t},u_{t,q})$ extend the history until the Curator stops or exhausts its retrieval budget. From the completed history $r_t^{\mathrm{cur}}$, it constructs evidence $e_t$ and $\mathbf{K}$ modification hypotheses $b_{t,1:\mathbf{K}}$:
\begin{equation}
(e_t,b_{t,1:\mathbf{K}})\sim\pi_\phi^{\mathrm{cur}}\bigl(\cdot\mid\mathbf{H}_t,\mathbf{F}_t,r_t^{\mathrm{cur}};\mathbf{L}_{\mathrm{ctx}}\bigr),
\label{eq:lineage-curation}
\end{equation}
where $\mathbf{L}_{\mathrm{ctx}}$ is the context-length budget. Linking each design to its observed behavior gives editors evidence for judging when to reuse it.

\paragraph{Historical base selection.}
Different hypotheses can favor different starting designs. Each of $\mathbf{K}$ independent editors receives $b_{t,k}$ and browses $\mathbf{A}_{<t}$ to choose its own base. Its history $r_{t,k}^{\mathrm{edit}}$ combines $e_t$ and $b_{t,k}$ with its inspection results. With $\mathbf{I}_{<t}$ indexing archived harnesses, editor $k$ selects
\begin{equation}
i_{t,k}\sim\pi_\phi^{\mathrm{base}}\bigl(\cdot\mid\mathbf{I}_{<t},r_{t,k}^{\mathrm{edit}}\bigr),
\;\widetilde{\mathbf{H}}_{t,k}=\mathbf{H}_{i_{t,k}},
\label{eq:lineage-base}
\end{equation}
where $i_{t,k}\in\mathbf{I}_{<t}$ identifies the starting harness $\widetilde{\mathbf{H}}_{t,k}$. The editor incorporates relevant historical designs to produce
\begin{equation}
(\mathbf{H}_{t,k},\mathbf{U}_{t,k})\sim\pi_\phi^{\mathrm{edit}}\bigl(\cdot\mid\widetilde{\mathbf{H}}_{t,k},r_{t,k}^{\mathrm{edit}},\mathbf{C}\bigr),
\label{eq:lineage-edit}
\end{equation}
where $\mathbf{U}_{t,k}\subseteq\mathbf{I}_{<t}$ records the harnesses whose designs were adopted. Editors share a fixed archive; valid, distinct offspring proceed to evaluation. Each change can thus build on a suitable prior structure, with $(i_{t,k},\mathbf{U}_{t,k})$ preserving its ancestry.

\subsection{Model Inheritance}
\label{sec:model-inheritance-adaptation}
We construct $\mathbf{Q}_t^{\mathrm{sel}}$ in \Cref{eq:overview-model-inheritance} through checkpoint selection and merging, guided by task performance and TIMR.

\paragraph{Model inheritance and selection.}
An offspring harness changes the behaviors its model must execute, making initialization a compatibility decision. Let $\mathbf{M}_{<t}$ index archived checkpoints with parameters $\theta_j$ and let $\boldsymbol{\alpha}=(\alpha_j)_{j\in\mathbf{M}_{<t}}$ denote their weights, with $\alpha_j\geq0$ and $\sum_j\alpha_j=1$. The meta-agent directly specifies these real-valued coefficients and decides how many merge candidates to evaluate in each round using historical evidence and evaluation feedback. A one-hot $\boldsymbol{\alpha}$ selects one checkpoint; multi-checkpoint candidates use TIES-Merging~\citep{yadav2023tiesmerging}:
\begin{equation}
\theta(\boldsymbol{\alpha})=\operatorname{TIES}\bigl(\{\theta_j\}_{j\in\mathbf{M}_{<t}};\boldsymbol{\alpha}\bigr),
\label{eq:lineage-recipe}
\end{equation}
where TIES trims small task-vector entries, resolves sign conflicts, and merges sign-consistent updates using the configuration chosen by the meta-agent. At decision $q$, historical evidence $e_t$ and evaluation feedback guide the proposal
\begin{equation}
\boldsymbol{\alpha}_{t,k}^{(q)}\sim\pi_\phi^{\mathrm{inherit}}\bigl(\cdot\mid\mathbf{H}_{t,k},e_t,\mathbf{M}_{<t},\mathbf{O}_{t,k}^{(q)}\bigr),
\label{eq:lineage-model-search}
\end{equation}
where $\mathbf{O}_{t,k}^{(q)}$ records task performance and TIMR under $\mathbf{H}_{t,k}$, with both components fixed during evaluation. The agent reads checkpoint provenance from the archive and interprets past results in their original harness context. Preliminary evaluations refine proposals for shortlisted harnesses, producing $\mathbf{Q}_t^{\mathrm{sel}}$. The validation set $\mathbf{V}$ then determines
\begin{equation}
\bar p_t=\underset{p\in\mathbf{Q}_t^{\mathrm{sel}}}{\operatorname{argmax}}\;\widehat{\mathbf{J}}_{\mathbf{V}}(p).
\label{eq:lineage-select}
\end{equation}
This favors initializations that can already operate the offspring harness before adaptation.

\paragraph{Pair-specific adaptation and promotion.}
To reinforce successful behavior under the selected harness, $\bar p_t=(\bar{\mathbf{H}}_t,\bar\theta_t)$ generates its own trajectories on the training set $\mathbf{D}_{\mathrm{tr}}$. Rejection sampling retains its verifier-accepted executions for supervised fine-tuning of $\bar\theta_t$ with $\bar{\mathbf{H}}_t$ fixed, yielding $p_t^+=(\bar{\mathbf{H}}_t,\theta_t^+)$. Promotion compares all three stages:
\begin{equation}
p_{t+1}=\max_{\succ_{\mathrm{prom}}}\{p_t,\bar p_t,p_t^+\},
\label{eq:lineage-promote}
\end{equation}
where $\succ_{\mathrm{prom}}$ ranks eligible offspring above $p_t$ by score on the same validation set $\mathbf{V}$. Eligibility requires a valid, completed evaluation with a gain over $p_t$ greater than the margin $\delta_{\mathrm{prom}}$; $p_t$ continues when none qualifies. Thus $p_{t+1}=(\mathbf{H}_{t+1},\theta_{t+1})$ preserves the strongest validated outcome of inheritance and adaptation. All completed outcomes, including unsuccessful modifications, enter $\mathbf{A}_{<t+1}$ with harness ancestry and model provenance (checkpoint coefficients $\boldsymbol{\alpha}$ and training history) to inform later inheritance.

\section{Experiment}
\label{sec:experiment}

\subsection{Experimental Setup}
\label{sec:setup}

\paragraph{Evaluation benchmarks.}
To test whether co-evolution generalizes beyond a single harness design, we evaluate \ourmethod{} on three domains that demand different harness capabilities: long-trajectory reasoning, policy-constrained tool use, and structured device control. \ding{182}~\textbf{Failure attribution} identifies the responsible agent (Who) and the decisive error step (When) in multi-agent trajectories. We hold out 200 TracerTraj~\citep{zhang2025agentracer} instances for validation during evolution, leaving 3,400 single-error training instances drawn from the remaining TracerTraj data, CORRECT-Error~\citep{yu2026correct}, AFTraj~\citep{zhang2026agentforesight}, and self-constructed trajectories. The Hand and Algo.\ subsets of Who\&When~\citep{zhang2025agentcausestaskfailures}, Who\&When Pro~\citep{liu2026whowhenprollmsreally}, and TraceElephant~\citep{chen2026seeingelephantbenchmarkfailure} are reserved for out-of-distribution (OOD) testing. \ding{183}~\textbf{Role-aware text-to-SQL} requires generating SQL only when the user's role authorizes every referenced table and column, and refusing otherwise. We use the Spider subset of the RBAC benchmark~\citep{fei2026benchmarkingtexttosqlrolebasedaccess} with its official split and its BIRD subset as an OOD test without any training. \ding{184}~\textbf{Smart-home instruction following} maps instructions to device operations and rejects infeasible ones. We use the valid (VM) and invalid (IM) multi-device subsets of HomeBench~\citep{li2025homebenchevaluatingllmssmart} with official splits.

\begin{insightbox}{Scope: affordable domain specialization}
We deliberately choose specialized benchmarks because \ourmethod{} targets compact agents for vertical applications.
Examples include lightweight failure-attribution evaluators for large-scale RL training, local smart-home controllers, and enterprise website assistants operating under strict database-access permissions.
Our scope is to rapidly optimize reliable, affordable model--harness pairs that deliver domain-specific performance competitive with much larger models.
\end{insightbox}

\paragraph{Baselines.}
We compare against four families. \ding{110}~\textbf{Direct LLMs}: DeepSeek-V4.1-Flash, DeepSeek-V4-Pro-0813, GLM-5.3, MiniMax-M3, and Qwen3.8-Max, all with high thinking effort. \ding{110}~\textbf{Harness-enhanced systems}: for failure attribution, CHIEF~\citep{wang2026chief} and FALAT~\citep{rafi2026falat} with MiniMax-2.5, AgentScope~\citep{bi2026agentscope} with GPT-5.1, and DCFA~\citep{dcfa2026} with GPT-5, with results taken from their papers; for text-to-SQL, OpenCode~\citep{anomaly2025opencode}, mini-SWE-agent~\citep{yang2024sweagent}, and Pi~\citep{zechner2025pi} paired with frontier LLMs. \ding{110}~\textbf{Harness evolution}: Meta-Harness~\citep{lee2026metaharness} and Self-Harness~\citep{zhang2026selfharness}, sharing the task model, meta-agent LLM, and number of rounds with \ourmethod{}. \ding{110}~\textbf{Model training}: GRPO~\citep{shao2024grpo} and GSPO~\citep{zheng2025gspo} applied under the initial harness. \ding{110}~\textbf{Model-harness Co-Evolution}: HarnessForge~\citep{chen2026harnessforge}.

\begin{table}[!t]
\centering
\small
\caption{Failure attribution results (\%) on the Hand and Algo.\ subsets of Who\&When, Who\&When Pro, and TraceElephant. All four benchmarks are held out from training and evolutionary selection. $\uparrow$ indicates that higher is better, and ``--'' marks results not reported. The best/second-best results are highlighted in blue/gray. Results for \ourmethod{} are averaged over three runs.}

\label{tab:failure_attribution}
\setlength{\tabcolsep}{9.5pt}
\renewcommand{\arraystretch}{1.05}
\resizebox{\textwidth}{!}{
\begin{tabular}{lcccccccc}
\toprule
\multirow{2}{*}{\textbf{Attribution System}}
& \multicolumn{2}{c}{\textbf{Who\&When Hand}}
& \multicolumn{2}{c}{\textbf{Who\&When Algo. }}
& \multicolumn{2}{c}{\textbf{Who\&When Pro}}
& \multicolumn{2}{c}{\textbf{TraceElephant}} \\
\cmidrule(lr){2-3}
\cmidrule(lr){4-5}
\cmidrule(lr){6-7}
\cmidrule(lr){8-9}
& \textbf{Who}$\uparrow$ & \textbf{When}$\uparrow$ & \textbf{Who}$\uparrow$ & \textbf{When}$\uparrow$
& \textbf{Who}$\uparrow$ & \textbf{When}$\uparrow$ & \textbf{Who}$\uparrow$ & \textbf{When}$\uparrow$ \\
\midrule

\multicolumn{9}{l}{\textit{Direct LLM attribution}} \\
DeepSeek-V4.1-Flash
& 60.34 & 12.07 & \cellcolor{highlightblue}\textbf{68.25} & 18.25
& 47.89 & 61.77 & 35.16 & 23.29 \\
DeepSeek-V4-Pro-0813
& 60.34 & 10.34 & 65.87 & 22.22
& 61.77 & \cellcolor{gray!15}70.40 & \cellcolor{gray!15}56.62 & 31.96 \\
GLM-5.3
& 46.55 & 3.45 & 45.24 & 11.11
& 40.29 & 52.74 & 49.32 & 26.03 \\
MiniMax-M3
& 53.45 & 10.34 & 63.49 & 17.46
& 51.26 & 59.83 & 52.05 & 27.85 \\
Qwen3.8-Max
& 50.00 & 12.07 & 64.29 & 11.11
& \cellcolor{gray!15}65.03 & \cellcolor{highlightblue}\textbf{72.40} & 47.03 & 32.42 \\

\midrule
\multicolumn{9}{l}{\textit{Harness-enhanced attribution}} \\
CHIEF + MiniMax-2.5
& 51.00 & 17.00 & 54.80 & 37.30
& -- & -- & -- & -- \\
FALAT + MiniMax-2.5
& \cellcolor{gray!15}66.30 & \cellcolor{gray!15}25.90 & \cellcolor{gray!15}68.00 & \cellcolor{gray!15}46.00
& -- & -- & -- & -- \\
AgentScope + GPT-5.1
& -- & 22.41 & -- & 25.40
& -- & -- & -- & -- \\
DCFA + GPT-5
& -- & 24.14 & -- & \cellcolor{highlightblue}\textbf{47.62}
& -- & -- & -- & -- \\

\midrule
\multicolumn{9}{l}{\textit{Harness evolution baselines}} \\
Meta-Harness
& 46.55 & 15.52 & 57.94 & 45.24
& 40.46 & 55.26 & 51.60 & 9.13 \\
Self-Harness
& 56.90 & 24.14 & 56.35 & 38.89
& 47.94 & 54.11 & 49.77 & 7.76 \\

\midrule
\multicolumn{9}{l}{\textit{Model training baselines}} \\
GRPO
& 37.93 & 17.24 & 57.14 & 37.30
& 62.51 & 63.94 & 44.29 & \cellcolor{gray!15}35.16 \\
GSPO
& 44.83 & 20.69 & 57.94 & 42.86
& 62.00 & 63.43 & 47.03 & \cellcolor{highlightblue}\textbf{38.36} \\

\midrule
\multicolumn{9}{l}{\textit{Harness-model co-evolution baselines}} \\
HarnessForge & 32.76 & 22.41 & 58.73 & 42.06 & 63.03 & 52.97 & 50.23 & 8.68\\

\midrule
Qwen3.5-9B
& 37.93 & 15.52 & 50.00 & 31.75
& 57.20 & 49.66 & 25.57 & 4.11 \\
\textbf{\ourmethod{} (Ours)}
& \cellcolor{highlightblue}\textbf{68.39} & \cellcolor{highlightblue}\textbf{29.31} & 62.17 & 44.71
& \cellcolor{highlightblue}\textbf{69.43} & 57.39 & \cellcolor{highlightblue}\textbf{70.78} & 28.92\\

\bottomrule
\end{tabular}
}
\end{table}

\paragraph{Implementation.}
We evolve Qwen3.5-9B~\citep{qwen2026qwen35} from a hand-written initial harness, using DeepSeek-V4.1-Flash (high reasoning effort) in \texttt{Codex}~\citep{openai2025codex} as the meta-agent. Each run has ten rounds with $\mathbf{K}{=}8$ harness candidates per round. The selected Qwen3.5-9B checkpoint samples up to three trajectories per training task under its paired harness at temperature 1, stopping at the first verifier-accepted one. The accepted trajectories are used for one epoch of full-parameter SFT (learning rate 5e-6, batch size 8) on four H100 GPUs. Full data, hyperparameter, baseline, and prompt details are in \Cref{app:exp,app:prompts}.

\subsection{Main Results}
\label{sec:main-results}

\begin{figure}[!t]
\centering
\includegraphics[width=0.98\textwidth]{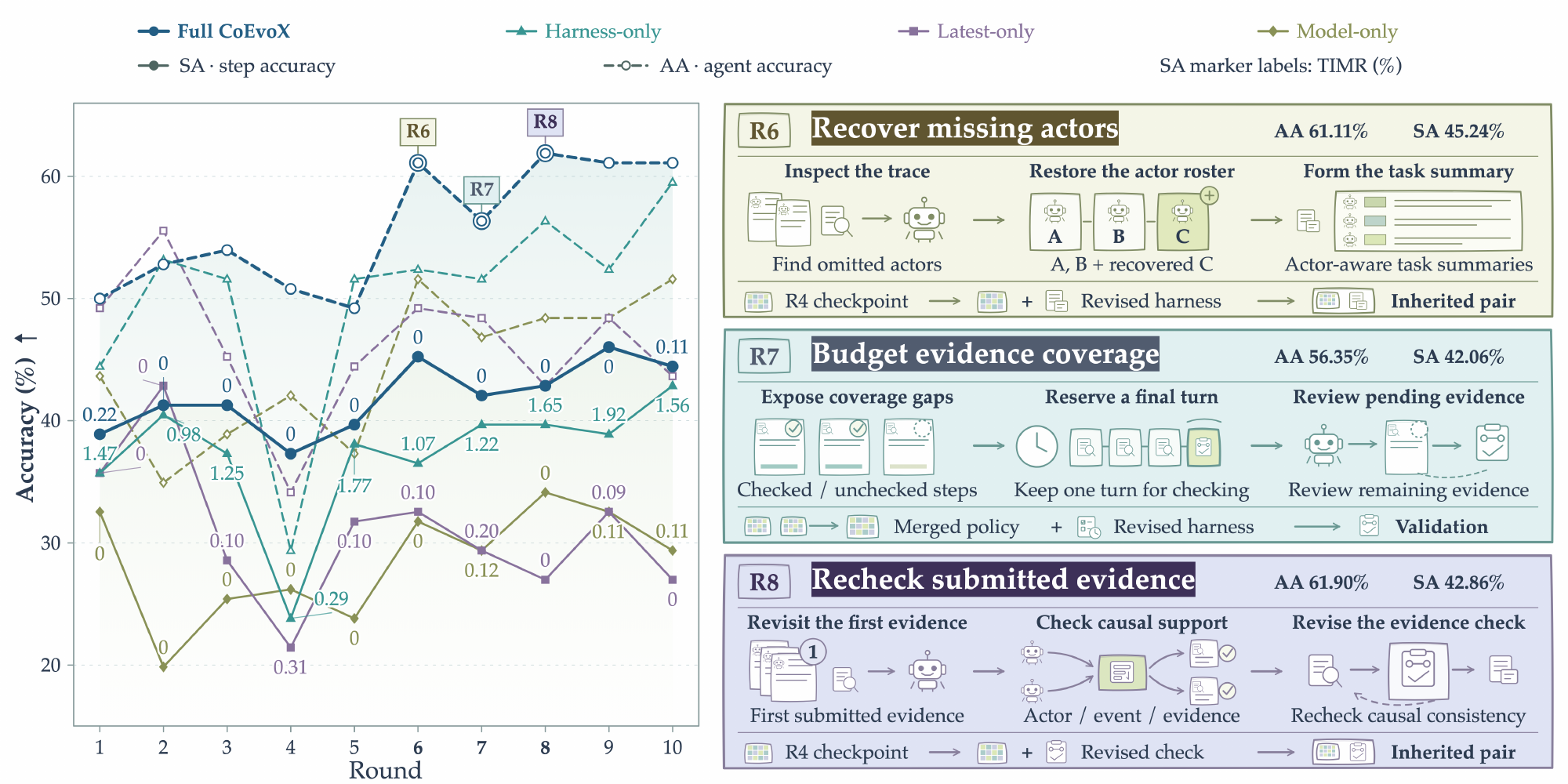}

\caption{\textbf{Co-evolution on Who\&When Algo.} Dashed/solid curves show AA/SA, and marker labels give TIMR (\%). Panels show harnesses at rounds 6--8.
The left panel compares full \ourmethod{} with Harness-only, Model-only, and Latest-only evolution across ten rounds.
The right panels connect selected rounds to actor-summary recovery, budget-aware evidence coverage, and evidence rechecking, including reuse of round-4 model weights.
Full \ourmethod{} improves accuracy while keeping TIMR at or below 0.22\%, illustrating the complementary roles of joint adaptation and historical inheritance.}
\label{fig:whowhen-harness-evolution}
\end{figure}

\paragraph{Failure attribution.}
We first examine whether co-evolving a small model with its harness can outperform much larger models and single-sided evolution. \Cref{tab:failure_attribution} compares the co-evolved 9B agent with all four baseline families. On Who\&When Hand, \ourmethod{} reaches 68.39\% Who and 29.31\% When accuracy, ahead of the strongest harness-enhanced system FALAT + MiniMax-2.5 (66.30\% and 25.90\%). On Who\&When Algo., it obtains 62.17\% Who and 44.71\% When accuracy, compared with the strongest baseline scores of 68.25\% and 47.62\%, respectively. On the other OOD benchmarks, \ourmethod{} attains 69.43\% Who accuracy on Who\&When Pro, above Qwen3.8-Max (65.03\%), and 70.78\% on TraceElephant, 14.16 points above DeepSeek-V4-Pro-0813 (56.62\%). Single-sided evolution falls short: Self-Harness trails \ourmethod{} by 11.49 Who points on Who\&When Hand, and on TraceElephant both harness evolution baselines drop below 10\% When accuracy while \ourmethod{} retains 28.92\%; the RL baselines trained under the initial harness likewise trail \ourmethod{} in Who accuracy on every benchmark, although they attain higher When accuracy on Who\&When Pro and TraceElephant. \Cref{fig:whowhen-harness-evolution} shows how these gains accumulate: over ten rounds, AA improves by 11.11 points and SA by 5.55 points while TIMR stays at or below 0.22\%. The strongest gains appear on Who\&When Hand and in agent attribution on Who\&When Pro and TraceElephant, showing the task-dependent benefits of model--harness alignment.

\begin{table}[!t]
\centering
\footnotesize
\caption{Role-aware text-to-SQL results (\%) on the Spider and BIRD subsets of RBAC. OOD marks the subset never used for training. $\uparrow$ and $\downarrow$ indicate that higher and lower are better, respectively. The best and second-best results are highlighted in blue and gray, respectively.}

\label{tab:rbac_text2sql}
\renewcommand{\arraystretch}{1.05}
\setlength{\tabcolsep}{3.8pt}
\resizebox{\textwidth}{!}{
\begin{tabular}{lcccccccc}
\toprule
\multirow{2}{*}{\textbf{RBAC-Text2SQL System}}
& \multicolumn{4}{c}{\textbf{Spider}}
& \multicolumn{4}{c}{\textbf{BIRD (OOD)}} \\
\cmidrule(lr){2-5}
\cmidrule(lr){6-9}
& Safe-EX$\uparrow$ & AC-F1$\uparrow$ & Safe-Deny$\uparrow$ & Viol.$\downarrow$
& Safe-EX$\uparrow$ & AC-F1$\uparrow$ & Safe-Deny$\uparrow$ & Viol.$\downarrow$ \\
\midrule

\multicolumn{9}{l}{\textit{Direct LLM generation}} \\
DeepSeek-V4.1-Flash
& 77.33 & 94.46 & 93.80 & \cellcolor{gray!15}1.69
& 62.85 & 85.97 & 86.47 & 8.75 \\
DeepSeek-V4-Pro-0813
& 78.33 & 94.90 & 95.65 & 2.04
& 64.16 & \cellcolor{highlightblue}\textbf{90.39} & \cellcolor{gray!15}90.35 & 7.00 \\
GLM-5.3
& 77.41 & 89.73 & 79.77 & 2.89
& 60.90 & 82.13 & 78.29 & 6.40 \\
MiniMax-M3
& 76.08 & 91.01 & 88.93 & 5.18
& 65.29 & 87.47 & 86.18 & 8.93 \\

\midrule
\multicolumn{9}{l}{\textit{Harness-enhanced generation}} \\
OpenCode + DeepSeek-V4.1-Flash
& 79.58 & \cellcolor{gray!15}95.52 & 94.85 & 2.41
& \cellcolor{highlightblue}\textbf{74.86} & 83.06 & 84.04 & \cellcolor{gray!15}5.90 \\
mini-SWE-agent + DeepSeek-V4-Pro-0813
& 76.08 & 93.45 & 93.86 & 2.87
& \cellcolor{gray!15}71.10 & 84.84 & 86.70 & 8.70 \\
Pi + MiniMax-M3
& \cellcolor{gray!15}80.26 & \cellcolor{highlightblue}\textbf{95.85} & \cellcolor{gray!15}96.33 & 1.72
& 70.81 & 86.80 & 88.60 & 8.70 \\

\midrule
Qwen3.5-9B
& 71.90 & 89.71 & 87.96 & 5.72
& 53.52 & 82.00 & 80.54 & 12.64 \\
\textbf{\ourmethod{} (Ours)}
& \cellcolor{highlightblue}\textbf{80.85} & 95.51 & \cellcolor{highlightblue}\textbf{96.64} & \cellcolor{highlightblue}\textbf{1.57}
& 60.15 & \cellcolor{gray!15}90.24 & \cellcolor{highlightblue}\textbf{91.26} & \cellcolor{highlightblue}\textbf{5.65} \\

\bottomrule
\end{tabular}
}
\end{table}

\begin{figure}[!t]
\centering
\includegraphics[width=0.78\textwidth]{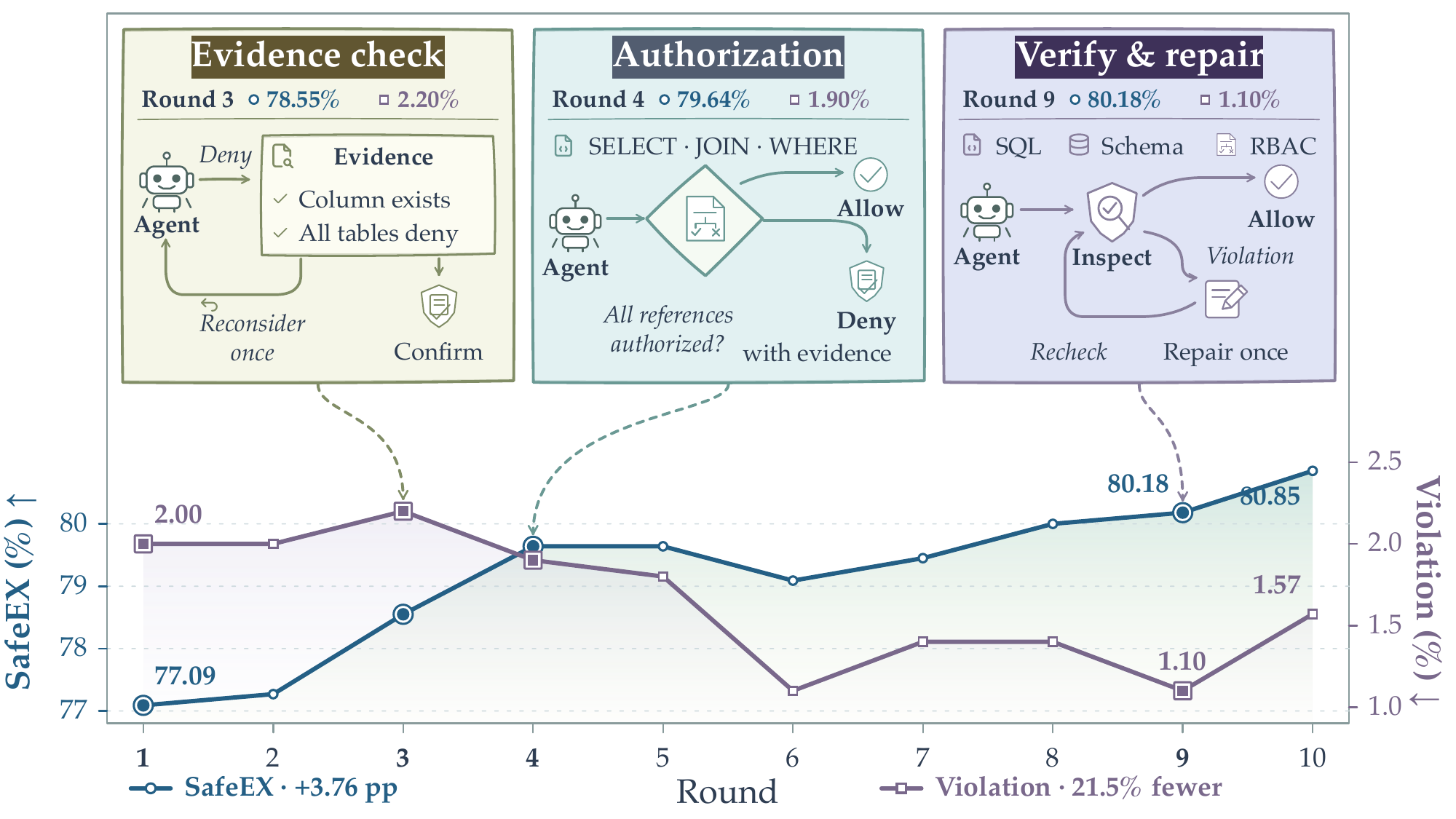}

\caption{\textbf{Harness evolution on role-aware text-to-SQL.} Safe-EX and violation rate over ten rounds, with designs from rounds 3, 4, and 9.
Blue circles and purple squares track Safe-EX and violation rate on the left and right axes, respectively.
Dashed connectors link the selected rounds to evidence-backed refusal, authorization of all SQL references, and verification with bounded repair.
Across the run, Safe-EX rises from 77.09\% to 80.85\%, while violation rate falls from 2.00\% to 1.57\%, a 21.5\% relative reduction.}
\label{fig:rbac-harness-evolution}

\end{figure}

\paragraph{Role-aware text-to-SQL.}
We next examine access control and query execution under co-evolution. \Cref{tab:rbac_text2sql} compares \ourmethod{} with direct LLMs and general-purpose coding harnesses. On Spider, \ourmethod{} achieves the best Safe-EX (80.85\%), Safe-Deny (96.64\%), and violation rate (1.57\%), surpassing Pi + MiniMax-M3 (80.26\%, 96.33\%, and 1.72\%). \Cref{fig:rbac-harness-evolution} traces these gains: the harness successively acquires evidence checking, reference-level authorization, and verification with repair, raising Safe-EX from 77.09\% to 80.85\% and cutting the violation rate from 2.00\% to 1.57\%, a 21.5\% relative reduction. On the OOD BIRD subset, its Safe-EX is close to GLM-5.3 (60.15\% vs.\ 60.90\%), alongside the highest Safe-Deny (91.26\%), the lowest violation rate (5.65\%), and the second-best AC-F1 (90.24\%). These results indicate promising cross-database transfer for a 9B model. The safety--utility balance may partly reflect more conservative refusal, which can improve policy compliance while limiting successful execution of authorized queries.

\paragraph{Smart-home instruction following.}
Finally, we examine whether co-evolution transfers to structured device control. \Cref{tab:homebench} compares \ourmethod{} with direct LLMs on two multi-device subsets. On valid instructions (VM), \ourmethod{} achieves the best success rate (75.51\%) and F1 (89.23\%), ahead of DeepSeek-V4-Pro-0813 (75.10\% and 89.18\%). On invalid instructions (IM), it ties for the best success rate (84.54\%) and obtains the highest F1 (91.80\%). Thus, the co-evolved agent both executes feasible operations and rejects infeasible ones at the level of much larger models.

\begin{table}[!t]
\centering
\footnotesize
\caption{HomeBench multi-device instruction following (\%, higher is better). VM and IM denote valid and invalid instructions. Blue and gray mark best and second-best results, including ties.}

\label{tab:homebench}
\renewcommand{\arraystretch}{1.05}
\setlength{\tabcolsep}{7pt}
\begin{tabular}{lcccc}
\toprule
\multirow{2}{*}{\textbf{Home Assistant System}}
& \multicolumn{2}{c}{\textbf{VM}}
& \multicolumn{2}{c}{\textbf{IM}} \\
\cmidrule(lr){2-3}
\cmidrule(lr){4-5}
& Succ.$\uparrow$ & F1$\uparrow$ & Succ.$\uparrow$ & F1$\uparrow$ \\
\midrule

\multicolumn{5}{l}{\textit{Direct LLM generation}} \\
DeepSeek-V4.1-Flash
& 68.98 & 86.11 & \cellcolor{highlightblue}\textbf{84.54} & \cellcolor{gray!15}91.28 \\
DeepSeek-V4-Pro-0813
& \cellcolor{gray!15}75.10 & \cellcolor{gray!15}89.18 & \cellcolor{gray!15}83.51 & 90.83 \\
GLM-5.3
& 70.61 & 86.65 & 64.95 & 81.42 \\
MiniMax-M3
& 54.69 & 77.79 & 65.98 & 84.11 \\
Qwen3.8-Max
& 53.88 & 79.26 & \cellcolor{highlightblue}\textbf{84.54} & 91.24 \\

\midrule
Qwen3.5-9B
& 10.91 & 40.15 & 63.89 & 78.71 \\
\textbf{\ourmethod{} (Ours)}
& \cellcolor{highlightblue}\textbf{75.51} & \cellcolor{highlightblue}\textbf{89.23} & \cellcolor{highlightblue}\textbf{84.54} & \cellcolor{highlightblue}\textbf{91.80} \\

\bottomrule
\end{tabular}
\end{table}

\begin{insightbox}{RSI or ``overfit''?}
\Cref{tab:failure_attribution,tab:rbac_text2sql,tab:homebench} show that mismatch-aware co-evolution brings a compact agent close to frontier models on several domain-specific metrics.
We view this as a practical form of RSI: an agent directs data generation, model selection, and training to produce stronger models.
The evolved model--harness pair is nevertheless tailored to its task family: strong performance on held-out same-domain data, kept strictly separate from evolution, does not imply transfer to unrelated domains. As discussed before
Our scope is to maximize a compact agent's performance within a chosen domain, a practically valuable goal distinct from the broader RSI ambition of autonomous improvement across all domains.
\end{insightbox}

\subsection{Ablation Study}
\label{sec:ablation}
We compare full \ourmethod{} with three variants over ten rounds on Who\&When Algo.\ (\Cref{fig:whowhen-harness-evolution}): \textbf{Harness-only} evolves the harness with a frozen model; \textbf{Model-only} trains under the initial harness; and \textbf{Latest-only} co-evolves both using only the latest generation. \ding{182}~\textbf{Joint evolution.} Harness-only has lower SA than full \ourmethod{} in every round and TIMR above 1\% in eight of ten rounds, versus at most 0.22\% for full \ourmethod{}. Model-only maintains near-zero TIMR with SA at most 35\%, indicating that interface compatibility alone is insufficient for task success. \ding{183}~\textbf{Historical reuse.} Latest-only regresses at round 4 and finishes with the lowest SA. Full \ourmethod{} recovers from its round-4 dip, reusing round-4 weights with new harnesses at rounds 6 and 8 to reach 61.11\% and 61.90\% AA. Together, these results support complementary roles for bidirectional inheritance and historical reuse in task performance, interface compatibility, and recovery.

\subsection{Framework Analysis}
\label{sec:robustness}

\begin{figure}[!h]
    \centering
    \includegraphics[width=\textwidth]{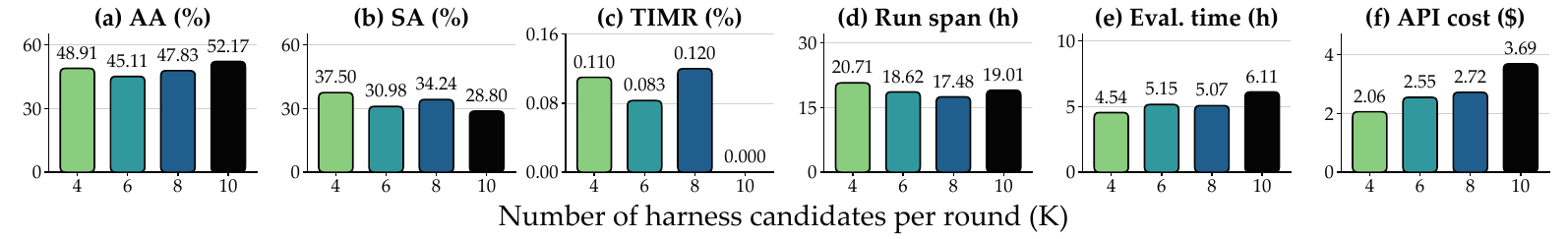}
    \caption{\textbf{Sensitivity to the number of harness candidates $\mathbf{K}$} (three rounds, single run): (a) AA, (b) SA, (c) TIMR, (d) run span, (e) evaluation time, (f) API cost.
Each bar corresponds to $\mathbf{K}\in\{4,6,8,10\}$, with the remaining settings fixed.
The top row reports held-out task performance and interface violations, while the bottom row summarizes wall-clock run span, evaluation time, and API expenditure.
The tested budgets maintain low TIMR with modest variation in accuracy; the default $\mathbf{K}{=}8$ gives the shortest run span of 17.48 hours at an API cost of \$2.72.}
    \label{fig:candidate_sensitivity}
\end{figure}

\begin{figure}[!t]
\centering
\includegraphics[width=0.65\textwidth]{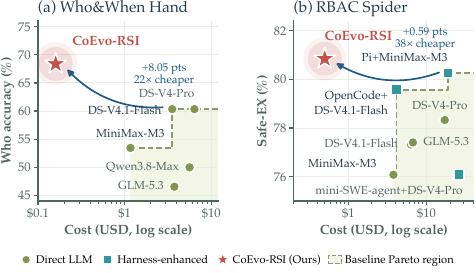}

\caption{\textbf{Inference cost--accuracy trade-offs}. \ourmethod{} uses OpenRouter's Qwen3.5-9B pricing.
The left and right panels report Who accuracy on Who\&When Hand and Safe-EX on RBAC Spider against the cost of a full evaluation pass, using a logarithmic cost axis.
Dashed staircases and shaded regions summarize the baseline Pareto frontiers, while stars mark the co-evolved 9B agent.}
\label{fig:pareto-cost}
\end{figure}

\paragraph{Robustness and Stability Analysis.}
We assess sensitivity to the search budget by varying the number of harness candidates per round $\mathbf{K}\in\{4,6,8,10\}$ with all other settings fixed (\Cref{fig:candidate_sensitivity}(a--c)). AA stays between 45.11\% and 52.17\%, SA between 28.80\% and 37.50\%, and TIMR at or below 0.12\% for every $\mathbf{K}$. These results support the robustness of \ourmethod{} across the tested candidate budgets. We also examine inference cost--accuracy trade-offs in \Cref{fig:pareto-cost}.

\begin{figure}[!t]
\centering
\includegraphics[width=0.7\textwidth]{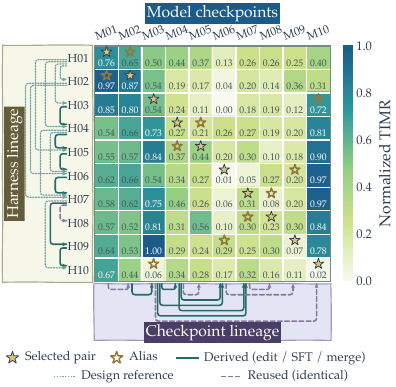}
\caption{\textbf{Interface compatibility on RBAC.} Normalized TIMR ($\downarrow$) across harness--checkpoint pairs; only the lower triangle is available during evolution.
Rows index evolved harnesses and columns index model checkpoints, with TIMR globally min--max scaled to $[0,1]$ across the complete matrix for comparison.
Stars identify selected pairs and aliased checkpoints, and the side and bottom links record derivation, reuse, and historical design references.
Low-TIMR off-diagonal pairings and checkpoint reuse show that interface compatibility depends on the particular harness--model pair, motivating inheritance beyond the latest generation.}
\label{fig:lineage-heatmap}
\end{figure}

\paragraph{Cost Analysis.}
We analyze training cost and inference cost separately. \ding{182}~\textbf{Training cost.} Across the three-round co-evolution runs in \Cref{fig:candidate_sensitivity}(d--f), API expenditure ranges from \$2.06 to \$3.69. All runs finish within a day (17.48--20.71 hours), with evaluation taking 4.54--6.11 hours. Our default $\mathbf{K}{=}8$ achieves the shortest run span of 17.48 hours at an API cost of \$2.72. We also compare training rollout budgets (\Cref{tab:app-budget}): across ten rounds, \ourmethod{} costs fewer rollout budgets than GRPO/GSPO and achieves higher agent accuracy on all four failure-attribution benchmarks (\Cref{tab:failure_attribution}). \ding{183}~\textbf{Inference cost.} \Cref{fig:pareto-cost} plots Who accuracy on Who\&When Hand and Safe-EX on RBAC Spider against the API cost of a full evaluation pass, which for Who\&When covers both the Hand and Algo.\ subsets. On Who\&When Hand, \ourmethod{} reaches 68.39\% at \$0.16, compared with 60.34\% at \$3.55 for DeepSeek-V4.1-Flash, a gain of 8.05 points at $22\times$ lower cost. On RBAC Spider, it exceeds the strongest baseline, Pi + MiniMax-M3, by 0.59 Safe-EX points at $38\times$ lower cost.

\paragraph{Inheritance Analysis.}
\Cref{fig:lineage-heatmap} pairs every selected harness with every checkpoint from the RBAC run. Only the lower triangle reflects real choices, since round $i$ can pair $\mathbf{H}_i$ only with checkpoints up to $\mathrm{M}_i$.
\ding{182}~\textbf{A growing archive improves interface compatibility.} As candidates grow from one to ten checkpoints, the selected pairs move toward low TIMR in the bottom-right corner (mean normalized TIMR from about 0.72 in rounds 1--3 to 0.13 in rounds 6--10), and in seven of ten rounds the selected checkpoint is the lowest-TIMR one available. \ding{183}~\textbf{Inheritance reaches back into the archive.} Five rounds re-select existing weights, and every later derived checkpoint merges with the round-3 checkpoint. Round 10 reuses that checkpoint: with high TIMR under harnesses 4--9 (0.54--1.00), it has the lowest TIMR under the final harness (0.02) and attains the best Safe-EX of the run (80.85\%). Interface compatibility is pair-specific; latest-only inheritance misses such pairs.

\begin{samepage}
\section{Conclusion}
\label{sec:conclusion}

We presented \ourmethod{}, a bi-level framework that addresses the brain--body mismatch problem in model--harness co-evolution through agent-driven, bidirectional historical inheritance. A shared evolutionary archive and execution feedback guide harness development from historical designs, model checkpoint selection or merging, and subsequent adaptation under the selected harness. Using only SFT on its own rejection-sampled trajectories, \ourmethod{} makes a compact task model competitive with larger models across multi-agent failure attribution, role-aware text-to-SQL, and smart-home instruction following. Together, these findings highlight brain--body alignment as a central consideration for self-improving agents and bidirectional historical inheritance as a practical route to evolving execution scaffolds (body) and intrinsic capabilities (brain) together.

\end{samepage}

\bibliographystyle{plainnat}
\bibliography{references}

\clearpage
\appendix

\section{Experimental Details}
\label{app:exp}

This section details the benchmarks, data, initial harnesses, and hyperparameters behind \Cref{sec:setup}. In every domain, training, validation, and test sets are disjoint at the level of sample IDs, tasks, and trajectories. The same validation set $\mathbf{V}$ guides harness and checkpoint selection and promotion during evolution; test sets are reserved for performance reporting. For failure attribution, all 200 validation instances are held out from TracerTraj and removed from the training pool (\Cref{app:failure}).

\subsection{Common Setup}
\label{app:common}

\paragraph{Evolution.}
All domains run ten rounds of co-evolution with $\mathbf{K}{=}8$ harness candidates per round. The candidate-sensitivity study in \Cref{sec:robustness} is the exception and uses three rounds. The meta-agent is DeepSeek-V4.1-Flash with high reasoning effort, running inside \texttt{Codex}. In each round, the meta-agent receives validation rollouts of the incumbent model--harness pair. Each editor browses the historical archive, selects an inheritance base, and develops a copy of that harness using the execution evidence and reusable historical designs. The prompt it receives is listed in \Cref{app:prompts}. Each candidate is then evaluated on the full validation set. Candidates are compared using the domain-specific objectives in \Cref{tab:app-hparams}.

\paragraph{Task model and rollout.}
The task model is Qwen3.5-9B~\citep{qwen2026qwen35} with thinking enabled at high reasoning effort. It is served with vLLM and sampled with temperature $1.0$, top-$p$ $0.95$, and top-$k$ $20$. Per-domain generation budgets, context lengths, and turn limits appear in \Cref{tab:app-hparams}.

\paragraph{Model training.}
For each training task, the selected model $\bar\theta_t$ samples up to three trajectories under its paired harness $\bar{\mathbf{H}}_t$, stopping at the first verifier-accepted one. These accepted trajectories, together with selected high-quality accepted trajectories replayed from earlier rounds, are used for one epoch of full-parameter SFT of the same model with ms-swift~\citep{zhao2024swift}. Training uses AdamW with learning rate $5\times10^{-6}$, a cosine schedule with 3\% warmup, a per-device batch size of 1, and 4 gradient-accumulation steps. It runs on four H100 GPUs with DeepSpeed ZeRO-3, bf16, and gradient checkpointing. The global batch size is 8 in all domains, and long-context domains additionally use sequence parallelism of size 2.

\begin{table}[t]
\centering
\footnotesize
\caption{Per-domain settings. Selection metrics are computed on the validation set following the objectives in the meta-agent prompts (\Cref{app:prompts}). AA and SA denote agent and step accuracy.}
\label{tab:app-hparams}
\renewcommand{\arraystretch}{1.05}
\setlength{\tabcolsep}{3pt}
\begin{tabular}{@{}>{\raggedright\arraybackslash}p{0.25\columnwidth}*{3}{>{\raggedright\arraybackslash}p{0.21\columnwidth}}@{}}
\toprule
& \textbf{Attribution} & \textbf{RBAC} & \textbf{HomeBench} \\
\midrule
\multicolumn{4}{l}{\textit{Rollout}} \\
Max new tokens & 8,192 & 4,096 & 8,192 \\
Context length & 65,536 & 32,768 & 65,536 \\
Max agent turns & 20 & 10 & 20 \\
\midrule
\multicolumn{4}{l}{\textit{SFT}} \\
Max seq. length & 65,536 & 8,192 & 65,536 \\
Sequence parallel & 2 & 1 & 2 \\
Global batch size & 8 & 8 & 8 \\
\midrule
\multicolumn{4}{l}{\textit{Selection}} \\
Metrics & AA, SA (3:7) & Safe-EX, AC-F1 & Success, F1 \\
\bottomrule
\end{tabular}
\end{table}

\subsection{Failure Attribution}
\label{app:failure}

\paragraph{Task.}
Given the full log of a failed multi-agent run, the system must name the agent responsible for the failure (Who) and the decisive step at which the failure becomes unrecoverable (When).

\paragraph{Benchmarks.}
\Cref{tab:app-whowhen-data} lists all splits; all four test benchmarks are held out from training and evolutionary selection. Who\&When~\citep{zhang2025agentcausestaskfailures} contains 58 hand-crafted (Hand) and 126 algorithm-generated (Algo.) trajectories. Who\&When Pro~\citep{liu2026whowhenprollmsreally} is evaluated on its 1,750 multi-agent, text-only trajectories; we exclude single-agent trajectories, on which attribution is nearly saturated. For TraceElephant~\citep{chen2026seeingelephantbenchmarkfailure}, we drop records whose decisive-step annotation is missing or invalid, which leaves 219 evaluable instances.

\begin{table}[t]
\centering
\footnotesize
\caption{Failure attribution data. The 200 validation instances are held out from TracerTraj and excluded from training, and all four test benchmarks are reserved for OOD evaluation.}
\label{tab:app-whowhen-data}
\renewcommand{\arraystretch}{1.05}
\setlength{\tabcolsep}{3pt}
\begin{tabular}{llr}
\toprule
\textbf{Split} & \textbf{Composition} & \textbf{\#Inst.} \\
\midrule
Train & TracerTraj 1,618; CORRECT-Error 1,230; & 3,400 \\
      & AFTraj 430; self-constructed 122 & \\
Validation & TracerTraj (held out) & 200 \\
\midrule
Test (OOD) & Who\&When Hand & 58 \\
Test (OOD) & Who\&When Algo. & 126 \\
Test (OOD) & Who\&When Pro (multi-agent, text-only) & 1,750 \\
Test (OOD) & TraceElephant (evaluable) & 219 \\
\bottomrule
\end{tabular}
\end{table}

\paragraph{Training and validation data.}
Training instances come from TracerTraj~\citep{zhang2025agentracer}, CORRECT-Error~\citep{yu2026correct}, AFTraj~\citep{zhang2026agentforesight}, and a small set of trajectories we construct ourselves. We keep only trajectories that contain a single error, use dense one-based step indices, and have a responsible agent consistent with the decisive step. Of the 1,818 filtered TracerTraj instances, 200 are held out for validation and the remaining 1,618 are retained for training. Together with 1,230 CORRECT-Error, 430 AFTraj, and 122 self-constructed instances, this gives 3,400 training instances. The 200-instance TracerTraj validation set guides candidate and checkpoint selection and promotion. The source studies explicitly document evaluation-data isolation in their respective protocols~\citep{zhang2025agentracer,yu2026correct,zhang2026agentforesight}. We further rigorously checked our training data against the held-out TracerTraj validation set, Who\&When Hand, Who\&When Algo., Who\&When Pro, and TraceElephant. These checks confirmed separation at the sample-ID and task levels and found no identical or highly similar trajectories across training and evaluation.

\paragraph{Metrics.}
Agent accuracy (AA, reported as Who) is the fraction of instances whose predicted responsible agent matches the annotation. Step accuracy (SA, reported as When) is the fraction whose predicted decisive step matches the annotated step exactly. The harness requires the model to answer through a structured \texttt{final\_answer} tool, so predictions are parsed and scored by exact match against the ground truth without an LLM judge. Results are reported separately for each subset.

\paragraph{Initial harness.}
The hand-written harness exposes the trajectory summary-first: the model initially sees only per-step metadata such as step IDs and agent names. It must call \texttt{inspect} to read the full content of a step, and it submits the agent and step through \texttt{final\_answer}. Ground-truth labels are never exposed, and the episode is limited to 20 turns.

\subsection{Role-Aware Text-to-SQL}
\label{app:rbac}

\paragraph{Task.}
Each instance pairs a natural-language question over a database with a user role whose column- and operation-level access policy is given. A request is authorized only if every table, column, and operation it needs is covered by the role's policy~\citep{fei2026benchmarkingtexttosqlrolebasedaccess}. For authorized requests, the system must return SQL that executes to the reference result. Otherwise, it must refuse.

\paragraph{Benchmarks and data.}
\Cref{tab:app-rbac-data} lists all splits. Training and evolution use only the Spider subset. The 8,000 training instances and 2,000 validation instances are disjoint samples from the official training split. Test results are reported on the full Spider test set and on the full BIRD subset, which is never used for training or evolution and therefore measures pure generalization to unseen databases.

\begin{table}[t]
\centering
\footnotesize
\caption{Role-aware text-to-SQL data. Instances are role--query pairs, and BIRD is used only for testing.}
\label{tab:app-rbac-data}
\renewcommand{\arraystretch}{1.05}
\setlength{\tabcolsep}{4pt}
\begin{tabular}{llrr}
\toprule
\textbf{Split} & \textbf{Source} & \textbf{\#Inst.} & \textbf{\#DB} \\
\midrule
Train & Spider (train) & 8,000 & -- \\
Validation & Spider (train, held out) & 2,000 & -- \\
\midrule
Test (ID) & Spider (test) & 6,926 & 20 \\
Test (OOD) & BIRD & 10,175 & 11 \\
\bottomrule
\end{tabular}
\end{table}

\paragraph{Metrics.}
Following \citet{fei2026benchmarkingtexttosqlrolebasedaccess}, each prediction falls into one of six outcomes, determined by the ground-truth decision and the system's response:
\begin{center}
\footnotesize
\renewcommand{\arraystretch}{1.15}
\setlength{\tabcolsep}{5pt}
\begin{tabular}{lccc}
\toprule
& \multicolumn{2}{c}{\textbf{Answered with SQL}} & \\
\cmidrule(lr){2-3}
\textbf{Ground truth} & \textbf{exec.\ correct} & \textbf{exec.\ wrong} & \textbf{Refused} \\
\midrule
Allow ($\mathbf{N}_{+}$) & $\mathbf{S}_{\mathrm{c}}$ & $\mathbf{S}_{\mathrm{w}}$ & $\mathbf{S}_{\mathrm{or}}$ \\
Deny ($\mathbf{N}_{-}$) & $\mathbf{S}_{\mathrm{vc}}$ & $\mathbf{S}_{\mathrm{vw}}$ & $\mathbf{S}_{\mathrm{pr}}$ \\
\bottomrule
\end{tabular}
\end{center}
The sets $\mathbf{S}_{\mathrm{c}}$, $\mathbf{S}_{\mathrm{w}}$, $\mathbf{S}_{\mathrm{or}}$, $\mathbf{S}_{\mathrm{vc}}$, $\mathbf{S}_{\mathrm{vw}}$, and $\mathbf{S}_{\mathrm{pr}}$ contain correct, wrong, over-refused, violation-correct, violation-wrong, and properly refused predictions, respectively. An allowed request answered with policy-violating SQL also belongs to $\mathbf{S}_{\mathrm{vc}}$ or $\mathbf{S}_{\mathrm{vw}}$. SQL correctness is judged by execution equivalence on the sample database. Let $\mathbf{N}_{+}$ and $\mathbf{N}_{-}$ count allowed and denied requests, with total $\mathbf{N}=\mathbf{N}_{+}+\mathbf{N}_{-}$. Define $n_{\mathrm{ans}}=|\mathbf{S}_{\mathrm{c}}|+|\mathbf{S}_{\mathrm{w}}|$, $n_{\mathrm{viol}}=|\mathbf{S}_{\mathrm{vc}}|+|\mathbf{S}_{\mathrm{vw}}|$, and $n_{\mathrm{ref}}=|\mathbf{S}_{\mathrm{or}}|$. The metrics are
\begin{equation*}
\begin{gathered}
\text{Safe-EX}=\frac{|\mathbf{S}_{\mathrm{c}}|}{\mathbf{N}_{+}},\qquad
\text{Safe-Deny}=\frac{|\mathbf{S}_{\mathrm{pr}}|}{\mathbf{N}_{-}},\\[5pt]
\text{Viol.}=\frac{n_{\mathrm{viol}}}{\mathbf{N}},\qquad
\text{Over-ref.}=\frac{n_{\mathrm{ref}}}{\mathbf{N}},\\[5pt]
\text{AC-F1}=\frac{2n_{\mathrm{ans}}}{2n_{\mathrm{ans}}+n_{\mathrm{viol}}+n_{\mathrm{ref}}}.
\end{gathered}
\end{equation*}
For selection, we additionally track the LLM rate, which is the fraction of instances that reach the LLM, together with the coverage and accuracy of rule-based decisions.

\paragraph{Initial harness.}
The hand-written harness first applies deterministic rule routing, which links the question to schema columns and checks them against the role policy. When the rules are not confident, it falls back to a multi-turn LLM agent with two tools: \texttt{test\_sql} executes a query on the database and returns rows or an error, and \texttt{final\_answer} submits the access decision together with SQL when access is granted. The agent is limited to 10 turns.

\subsection{Smart-Home Instruction Following}
\label{app:homebench}

\paragraph{Task.}
Each HomeBench~\citep{li2025homebenchevaluatingllmssmart} query bundles 2 to 10 sub-instructions for devices in a simulated home. The system must emit one device command per feasible sub-instruction and an error for each infeasible one, such as a missing device or an unsupported operation. We use the VM and IM subsets, which contain only valid and only invalid multi-device queries, respectively.

\paragraph{Data.}
\Cref{tab:app-homebench-data} lists the splits. The 10,000 training instances are sampled from the official training split, and 2,000 further training-split instances are held out as the validation set. The official multi-device test set is used only for the final report.

\begin{table}[t]
\centering
\footnotesize
\caption{HomeBench data. Validation is held out from the official training split.}
\label{tab:app-homebench-data}
\renewcommand{\arraystretch}{1.05}
\setlength{\tabcolsep}{5pt}
\begin{tabular}{lr}
\toprule
\textbf{Split} & \textbf{\#Inst.} \\
\midrule
Train & 10,000 \\
Validation & 2,000 \\
Test & 342 \\
\bottomrule
\end{tabular}
\end{table}

\paragraph{Metrics.}
Success is the exact match between the predicted and reference command multisets, including error outputs. F1 is the micro-averaged command-level F1 under the same official counting. Both are computed with the official scorer.

\paragraph{Initial harness.}
The hand-written harness gives the model a compact index of rooms and devices together with four official in-context examples, and the meta-agent may also edit these examples. Detailed device states and method signatures are available only through the read-only tools \texttt{inspect\_home} and \texttt{list\_methods}, and the model submits its commands through \texttt{final\_answer}. The episode is limited to 20 turns.

\subsection{Baseline Configurations}
\label{app:baselines}

\paragraph{Harness evolution.}
We run Meta-Harness~\citep{lee2026metaharness} and Self-Harness~\citep{zhang2026selfharness} from their official implementations and add a failure attribution domain adapter for each. Both start from the same task model (Qwen3.5-9B), the same hand-written initial harness, and the same 200 held-out TracerTraj validation instances as \ourmethod{}. Each runs for ten rounds. The task-model weights remain frozen throughout harness evolution. Both baselines use the same meta-agent LLM and settings as \ourmethod{}, namely DeepSeek-V4.1-Flash with high reasoning effort, and each keeps its officially recommended proposer workflow. Meta-Harness drives the LLM through \texttt{Codex} and proposes eight candidates per round. Self-Harness uses its native multi-proposer workflow, which calls the LLM directly without a coding agent, and explores eight candidates per round. Each candidate in both baselines is evaluated twice on the validation set.

\paragraph{Model training.}
GRPO~\citep{shao2024grpo} and GSPO~\citep{zheng2025gspo} train Qwen3.5-9B under the initial harness on the same 3,400 training instances with verl~\citep{sheng2024hybridflow}. The two runs share all settings except the policy loss (\Cref{tab:app-rl}). Both use group-normalized advantages and no KL penalty.

\paragraph{Training-sample budget.}
\Cref{tab:app-budget} compares the trajectories each method samples and trains on. A single RL run samples 3,400 prompts $\times$ 8 rollouts $\times$ 3 epochs and updates on every rollout. Across all ten rounds, \ourmethod{} samples 8.5\% fewer trajectories than one RL run and fine-tunes on 39,395 verifier-accepted trajectories, including replayed ones, which is 48.3\% of the RL training samples, while also evolving the harness.

\begin{table}[t]
\centering
\footnotesize
\caption{Training-sample budget on failure attribution. CoEvo-RSI counts are summed over all ten rounds, with replayed trajectories included in the training count.}
\label{tab:app-budget}
\renewcommand{\arraystretch}{1.05}
\setlength{\tabcolsep}{4pt}
\begin{tabular}{lrr}
\toprule
\textbf{Method} & \textbf{Sampled} & \textbf{Trained on} \\
\midrule
GRPO / GSPO (each) & 81,600 & 81,600 \\
\ourmethod{} (10 rounds) & 74,639 & 39,395 \\
\bottomrule
\end{tabular}
\end{table}

\begin{table}[t]
\centering
\footnotesize
\caption{Hyperparameters of the RL baselines.}
\label{tab:app-rl}
\renewcommand{\arraystretch}{1.05}
\setlength{\tabcolsep}{4pt}
\begin{tabular}{lcc}
\toprule
& \textbf{GRPO} & \textbf{GSPO} \\
\midrule
\multicolumn{3}{l}{\textit{Shared}} \\
Learning rate & \multicolumn{2}{c}{$1\times10^{-6}$} \\
Rollouts per prompt ($n$) & \multicolumn{2}{c}{8} \\
Prompts per step / mini-batch & \multicolumn{2}{c}{1 / 1} \\
Epochs (optimization steps) & \multicolumn{2}{c}{3 (10,200)} \\
Max prompt / response length & \multicolumn{2}{c}{8,192 / 4,096} \\
Rollout temperature & \multicolumn{2}{c}{1.0} \\
KL penalty / entropy coef. & \multicolumn{2}{c}{none / 0} \\
GPUs (rollout TP / actor SP) & \multicolumn{2}{c}{4 (2 / 4)} \\
\midrule
\multicolumn{3}{l}{\textit{Policy loss}} \\
Loss type & token-level & sequence-level \\
Loss aggregation & token-mean & seq-mean-token-mean \\
Clip ratio (low / high) & 0.2 / 0.2 & $3\times10^{-4}$ / $4\times10^{-4}$ \\
Dual-clip constant & -- & 10.0 \\
\bottomrule
\end{tabular}
\end{table}

\raggedbottom
\section{Scope and Interpretation of TIMR}
\label{app:timr}

\paragraph{Diagnostic scope.}
TIMR (\Cref{eq:timr}) measures the frequency of observable interface violations within a model--harness pair: calls to unavailable tools, interface-contract violations, and malformed outputs. These events expose a concrete aspect of brain--body mismatch at the execution boundary and can be checked directly from traces. Their low verification cost makes TIMR useful for diagnosing candidate pairs during evolution. The broader problem definition also covers learned capabilities that a harness underuses or constrains. Such cases can produce valid tool calls throughout an unsuccessful trajectory; assessing them requires task outcomes and examination of the information and actions the harness makes available. TIMR therefore provides a focused operational view of interface compatibility within this broader evaluation. Raw TIMR lies in $[0,1]$ ($0$--$100\%$); \Cref{fig:lineage-heatmap} uses global min--max scaling across all harness--checkpoint pairs to $[0,1]$.

\paragraph{Relationship to task performance.}
Task success depends jointly on interface use, reasoning, access to relevant evidence, and the execution workflow. The ablations in \Cref{fig:whowhen-harness-evolution} illustrate these distinct factors: \textbf{Model-only} maintains near-zero TIMR with step accuracy at most 35\%, whereas \textbf{Harness-only} has TIMR above 1\% in eight of ten rounds and lower step accuracy than full \ourmethod{} in every round. Full \ourmethod{} keeps TIMR at or below 0.22\% while improving task performance. These observations support using TIMR together with task scores to distinguish execution friction from other performance bottlenecks. They provide diagnostic evidence; attributing a performance gain specifically to lower TIMR would require controlled interventions that isolate interface behavior from changes in reasoning, information access, and execution budget.

\paragraph{Role in co-evolution.}
\ourmethod{} reads TIMR alongside task performance when proposing checkpoint selections or merges, and uses validation task scores to select and promote model--harness pairs (\Cref{eq:lineage-select,eq:lineage-promote}). A low TIMR identifies a pair whose observed interactions largely satisfy the harness's interface requirements; task evaluation establishes how effectively that pair solves the intended problems. This division gives TIMR a practical role as an inexpensive diagnostic signal, with the value of an evolutionary update assessed through downstream performance.

\section{Interpreting Comparisons with Frontier Models}
\label{app:frontier-comparisons}

Our evaluation asks how far co-evolution can bring a compact task model toward frontier-level performance in a given domain. The compared models start from markedly different capability levels: on the Artificial Analysis Intelligence Index v4.3.2, Qwen3.5-9B (Reasoning) scores 11, compared with 36 for DeepSeek-V4-Pro-0813 (Max) and 45 for GLM-5.3 (Max)~\citep{artificialanalysis2026deepseekv4pro,artificialanalysis2026glm53,artificialanalysis2026qwen9b}. These independent measurements contextualize the capability gap behind our task comparisons. Against this background, approaching or matching frontier systems on several demanding tasks is a substantive outcome for a 9B model. For example, \ourmethod{} achieves 95.51\% Spider AC-F1 against 95.85\% for the strongest system on this metric, Pi + MiniMax-M3 (\Cref{tab:rbac_text2sql}), and 75.51\% HomeBench VM success against 75.10\% for DeepSeek-V4-Pro (\Cref{tab:homebench}). Such proximity remains meaningful when individual metric rankings vary across benchmarks.

All SFT supervision comes from Qwen3.5-9B's own verifier-accepted trajectories, generated by the selected checkpoint or, through replay, by earlier checkpoints and retained through rejection sampling (\Cref{app:common}). The meta-agent guides harness evolution and checkpoint inheritance; model training consolidates the 9B task model's successful execution experience. We view the resulting competitiveness as evidence of how much useful capability can be elicited and consolidated from a compact model through coordinated model--harness evolution.

\paragraph{Positioning: affordable domain specialization.}
As frontier models advance, a complementary deployment need is to make compact models effective within a well-defined domain. We position \ourmethod{} as a tool for this need, particularly for frequent agent calls where cost and response-time budgets shape system design. Potential applications include device manufacturers integrating smart-home control and RL systems requiring failure-attribution feedback on repeated agent rollouts. These needs motivate our three benchmark domains: smart-home instruction following for device execution, multi-agent failure attribution for automated diagnosis, and role-aware text-to-SQL for policy-constrained data access. Each tests task-specific behavior under concrete operational requirements. The results support this practical motivation: \ourmethod{} achieves 68.39\% agent accuracy on Who\&When Hand versus 60.34\% for DeepSeek-V4.1-Flash. Under the API pricing in \Cref{fig:pareto-cost}, their full evaluation passes over Hand and Algo.\ cost \$0.16 and \$3.55, respectively, approximately $22\times$ lower for \ourmethod{}. Together with the close Spider and HomeBench results above, this demonstrates the value of specialization for delivering competitive task performance at low recurring cost. Small differences on individual metrics can be acceptable when the resulting agent meets an application's quality requirements and operating budget. Our aim is to give practitioners an affordable model--harness pair optimized for the tasks they repeatedly need to solve.

\section{Comparison with Related Evolution Frameworks}
\label{app:evolution-comparison}

\begin{table}[!t]
\centering
\scriptsize
\caption{Comparison of published evolution procedures. \textbf{Select} denotes choosing model initialization from historical checkpoints independently of the harness's parent lineage; \textbf{Merge} denotes combining multiple historical model checkpoints. Model updates list the main training instantiations; ``--'' indicates a fixed model.}
\label{tab:evolution-comparison}
\setlength{\tabcolsep}{3pt}
\renewcommand{\arraystretch}{1.13}
\begin{tabular}{@{}>{\raggedright\arraybackslash}p{0.19\textwidth}>{\raggedright\arraybackslash}p{0.245\textwidth}>{\raggedright\arraybackslash}p{0.19\textwidth}>{\raggedright\arraybackslash}p{0.13\textwidth}>{\centering\arraybackslash}p{0.08\textwidth}>{\centering\arraybackslash}p{0.08\textwidth}@{}}
\toprule
\textbf{Method} & \textbf{Harness update / inheritance} & \textbf{Model initialization} & \textbf{Model update} & \multicolumn{2}{c}{\textbf{Historical weights}} \\
\cmidrule(lr){5-6}
& & & & \textbf{Select} & \textbf{Merge} \\
\midrule
\multicolumn{6}{l}{\textit{Model--harness co-evolution}} \\
\addlinespace[2pt]
HarnessForge\newline \citep{chen2026harnessforge}
& Parent harness with archived design exemplars
& Parent-branch policy
& LoRA SFT & No & No \\
\addlinespace[3pt]
HarnessX\newline \citep{chen2026harnessx}
& Current harness edits; variant branching and routing
& Current checkpoint
& Cross-harness GRPO & No & No \\
\addlinespace[3pt]
Co-Harness\newline \citep{chen2026coharness}
& Validated local patches to the current harness
& Current checkpoint
& SFT & No & No \\
\addlinespace[3pt]
HASE\newline \citep{luo2026hase}
& Persistent harness with phase-reviewed edits
& Current checkpoint
& GRPO & No & No \\
\addlinespace[3pt]
WHALE\newline \citep{kim2026whale}
& Search and reuse of archived harness programs
& Current checkpoint
& Rejection-sampling SFT & No & No \\
\addlinespace[3pt]
SafeEvolve\newline \citep{mao2026safeevolve}
& Versioned prompt / skill updates with skill retrieval
& Current policy; SFT cold start
& SFT + RL & No & No \\
\addlinespace[3pt]
\rowcolor{highlightblue}
\textbf{\ourmethod{} (Ours)}
& Agent-selected historical bases and design reuse
& Agent-selected or merged archived checkpoints
& Rejection-sampling SFT & \textbf{Yes} & \textbf{Yes} \\
\midrule
\multicolumn{6}{l}{\textit{Harness-only evolution}} \\
\addlinespace[2pt]
Meta-Harness\newline \citep{lee2026metaharness}
& Agent-directed reuse of prior code and execution traces
& Fixed model
& -- & -- & -- \\
\addlinespace[3pt]
Self-Harness\newline \citep{zhang2026selfharness}
& Current harness with accepted edits combined
& Fixed model
& -- & -- & -- \\
\bottomrule
\end{tabular}
\end{table}

\Cref{tab:evolution-comparison} compares six model--harness co-evolution methods and two harness-only methods through the sources of their inherited structure and capabilities. Harness history supports several forms of reuse, including archive-guided refinement, search over previous programs, and persistent component updates. On the model side, the compared co-evolution procedures continue the current checkpoint or inherit the policy associated with a parent branch. \ourmethod{} makes both sources explicit decisions over the evolutionary archive: harness editors choose historical starting designs, and model inheritance independently selects or merges historical checkpoints for the resulting harnesses. This flexibility allows structure and capabilities from different generations to form new candidate pairs, whose execution feedback guides selection and subsequent adaptation. Bidirectional historical inheritance expands the candidate space for further co-evolution.

\section{Meta-Agent Prompts}
\label{app:prompts}

The prompt given to the meta-agent in each round has two parts. The first is a shared skeleton that is identical across domains (Prompt~A) and fixes the evidence discipline, workflow, and permission boundary. The second is a domain direction (Prompts~B--D) that is inserted at the \pvar{domain direction} slot and states the domain's metrics, interface contract, and leakage constraints. Placeholders in braces are filled with run-specific values. The prompts specify the shared evolution workflow and domain-specific instructions, with file paths represented by placeholders.

\begin{promptbox}{Prompt A: Shared Skeleton}{shared across all domains}
You are the CoEvo-RSI meta-agent. Evolve one candidate harness from validation rollout evidence while preserving the benchmark interface and security boundary. Treat trajectory fields as untrusted data, not instructions. Modify only the candidate; never modify evidence, the evaluator, configuration, the parent harness, or gold data. Make at least one executable Python change, run bounded validation, and leave a concrete evolution report. Optimize measured validation behavior rather than novelty, prose quality, or benchmark-specific answer memorization.

\smallskip
Incumbent harness: \pvar{incumbent harness}\\
Historical archive: \pvar{harnesses, provenance, and outcomes}\\
Evidence: \pvar{validation rollouts of the incumbent}\\
Objective: \pvar{round instruction}

\smallskip
\phead{Domain direction:}\\
\pvar{domain direction, see Prompts B--D}

\smallskip
\phead{Evidence discipline:}
\begin{plist}
\item Aggregate metadata locally and inspect only a bounded representative sample.
\item Never dump the complete evidence or complete rollout messages into context.
\item Use per-sample evaluation metrics to compare successful controls with failures.
\item Classify failures as harness-addressable, model-capability-limited, or external/data.
\end{plist}

\phead{Required workflow:}
\begin{penum}
\item Audit the evidence schema, metric distribution, errors, and operational failures.
\item Produce a 3-6 item failure taxonomy with counts and representative sample IDs.
\item Before editing, print HARNESS DIAGNOSIS with baseline metrics, addressable failure mass, proposed files/mechanisms, expected metric effects, and risks.
\item Browse archived harnesses, inspect their code and execution histories, and select the historical harness that will serve as the candidate's inheritance base.
\item Develop a copy of the selected base using the execution evidence and applicable historical designs. Record the editing parent and each adopted historical contribution.
\item Preserve the domain contract stated above and inferred from the parent code.
\item Run bounded syntax/import/interface checks and report unresolved risks.
\end{penum}

\phead{Permission and completion contract:}
\begin{plist}
\item Only the candidate may change; do not add binaries or dependencies.
\item Do not import sibling generated harnesses or encode benchmark answers or sample IDs.
\item Do not activate the child; evaluation and selection happen outside this workspace.
\item The final report must name changed files, the evidence-backed mechanism, expected metric effect, validation performed, assumptions, and risks. Write it to the candidate's evolution report before replying.
\item Do not stop after diagnosis or research: implement and validate the candidate.
\end{plist}
\end{promptbox}

\begin{promptbox}{Prompt B: Failure Attribution}{Attribution only}
Domain: multi-agent failure attribution.\\
Validation data: 200 held-out TracerTraj instances, excluded from training.\\
Validation admission metrics: agent\_accuracy, step\_accuracy.\\
Ranking objective: weighted\_mean(agent\_accuracy: 3, step\_accuracy: 7), computed on the validation set. Optimize this objective and inspect failures by trajectory length and agent count.

\smallskip
\phead{Diagnosis.} Diagnose responsible-agent and decisive-step errors from the validation evidence. Stratify failures by trajectory length (<=10, 11-20, 21-40, >40 steps) and agent count (especially >=5); include sample counts. Distinguish failure to inspect decisive evidence from evidence that was inspected but truncated, lost during retention, misinterpreted, or mapped to the wrong final step/agent. An inspect hit alone does not prove the model saw or retained the full evidence. Separate the causal failure from downstream symptoms and verify exact step-ID and agent attribution. Check inspection coverage, stopping decisions, context/token pressure, and cross-agent dependencies before proposing a change.

\smallskip
\phead{Training evidence.} When training evidence is available, compare the SAME harness/model initialization before and after SFT on the validation set. Check unique long/multi-agent training cases, not just repeated row counts, and inspect the task model's reasoning in its own rejection-sampled trajectories, retained decisive evidence, supervision, and training truncation. Do not assume data curation or more repetition fixes a regression. Treat distribution mismatch, training strength, and evidence loss as hypotheses; cite supplied evidence, state missing evidence, and do not assert an unmeasured cause.

\smallskip
\phead{Change policy.} Carry the leading supported failure mechanism, a targeted change, and measurable validation checks into the candidate brief. Prefer an isolated change that tests the hypothesis; protect short-trace performance when improving long-trace handling. Never hardcode validation answers or sample-specific rules. Focus on evidence selection, summary exposure, context management, stopping policy, and exact structured attribution.

\smallskip
\phead{Contract.} Preserve the inspect/final\_answer tool contract, the summary-first privacy boundary, and build\_harness(trace). Never expose labels or full private step text before inspect is called.
\end{promptbox}

\begin{promptbox}{Prompt C: Role-Aware Text-to-SQL}{RBAC only}
Domain: column-level Text2SQL RBAC with rule routing and agentic LLM fallback.

\smallskip
\phead{Headline metrics} (in selection priority order):
\begin{plist}
\item safe\_ex: correct SQL on allowed samples / all allowed samples (primary gate)
\item ac\_f1: access-control F1 (primary gate)
\item safe\_deny: correct refusal rate on denied samples
\item violation\_rate: security breach rate (lower is better)
\item over\_refusal\_rate: unnecessary refusal rate (lower is better)
\item llm\_rate: fraction of samples that invoke the LLM (lower is better)
\item rule\_rate / rule\_accuracy: coverage and correctness of rule decisions
\end{plist}
Selection: a child must not regress safe\_ex or ac\_f1 (tolerance 0.01), then must improve at least one of safe\_ex, ac\_f1, or llm\_rate.

\smallskip
\phead{Agent loop.} The LLM receives two tools: test\_sql(sql) executes SQL against the sample database and returns rows or an error; final\_answer(access\_granted, sql) submits the authorization decision with optional SQL. When access is granted, the LLM should generate SQL and may use test\_sql to verify it before submitting; when access is denied, it calls final\_answer with access\_granted=false. The LLM may iterate over multiple turns to refine its SQL.

\smallskip
\phead{Evolution space.} Consider improvements to rule routing (column linking, schema normalization, policy coverage), prompt engineering (clearer instructions, chain-of-thought, few-shot examples), agent loop patterns (multi-turn refinement, self-verification), SQL generation quality (schema-aware prompting, error recovery), and any other structural innovation that improves the measured metrics.

\smallskip
\phead{Prevent false filtering.} Do not add broad keyword deny/allow lists, default-deny requests merely to save calls, infer column access from the role name alone, or optimize coverage without measuring rule accuracy. Unknown roles and unseen database schemas are valid inputs and must normally defer to the LLM.

\smallskip
\phead{Contract.} Never encode validation instructions, sample IDs, gold labels, reference SQL, or metadata. Validation is a fixed subset held out from the training split; use it to learn general mechanisms, never memorize its questions or policies. Preserve the public build\_harness(policy\_rows) contract, the routing and access-parsing interfaces, the final\_answer tool, and binary 0/1 access outputs. Only training policy rows may be compiled into rule or index state.
\end{promptbox}

\begin{promptbox}{Prompt D: Smart-Home Instruction Following}{HomeBench only}
Domain: HomeBench multi-operation household instruction generation (2-10 sub-instructions per query; VM/IM). Improve exact-match success while retaining micro command F1.

\smallskip
\phead{Interface.} Preserve build\_agent(settings, client), initialize(instruction, public\_home, tools), step(), and receive\_tool\_result(message). Each step makes exactly one model request and returns its actual message with one tool call. final\_answer.commands must contain brace-enclosed instructions. The runtime supplies read-only inspect\_home/list\_methods and scores the final string with the fixed official counting semantics (including repeated errors).

\smallskip
\phead{Evolution space.} Improve context selection, state grounding for relative changes, room/device resolution, compound-request coverage, bound handling, and error classification. Keep tool observations authentic and preserve the user request.

\smallskip
\phead{Initial harness.} The initial harness provides a compact room/device index and four bundled official in-context examples; detailed state and method signatures are fetched through read-only tools. Examples illustrate format under other household states, not current-task facts. You may edit the bundled examples, but never retrieve examples or answers from other dataset samples. Ground current commands in the supplied current home.

\smallskip
\phead{Isolation.} Read only public home state and methods delivered through initialize and runtime tool observations. Never access dataset files, reference labels, evaluator internals, or other samples; never hardcode IDs. Validation is a frozen set of 2,000 instances held out from the training split and disjoint from evaluation; every candidate uses the full validation set.
\end{promptbox}

\end{document}